\documentclass{article}

\usepackage{amsmath,amsfonts,bm}

\def\eqref#1{equation~\ref{#1}}

\def\1{\bm{1}}

\def\rb{{\textnormal{b}}}

\def\vb{{\bm{b}}}

\DeclareMathAlphabet{\mathsfit}{\encodingdefault}{\sfdefault}{m}{sl}
\SetMathAlphabet{\mathsfit}{bold}{\encodingdefault}{\sfdefault}{bx}{n}

\let\ab\allowbreak

\pdfoutput=1

\usepackage{arxiv}

\usepackage[utf8]{inputenc} 

\usepackage{url}            
\usepackage{booktabs}       
\usepackage{amsfonts}       
\usepackage{nicefrac}       
\usepackage{lipsum}		
\usepackage{graphicx}
\usepackage{natbib}
\usepackage{doi}

\usepackage{amsmath}
\usepackage{wrapfig}
\usepackage{Definitions}
\usepackage{makecell}
\usepackage{bbm} 
\usepackage{enumerate}
\usepackage[ruled,vlined]{algorithm2e}
\usepackage{algorithmic}
\usepackage{tabularx}
\usepackage{adjustbox}
\usepackage[T1]{fontenc}    
\usepackage{comment}
\usepackage{color}
\usepackage{multirow}
\usepackage{multicol}
\usepackage{float}
\usepackage{pdfpages}
\usepackage{microtype}      
\usepackage{xcolor}         
\usepackage{mathtools}
\usepackage{empheq}

\usepackage{blindtext}
\usepackage{bm}
\usepackage{diagbox}

\title{Hawkes Processes with Delayed Granger Causality}
\date{ } 

\author{{Chao Yang} \\
	School of Data Science \\
	The Chinese University of Hong Kong, Shenzhen\\
	Shenzhen 518172, China \\
	\texttt{222043011@link.cuhk.edu.cn} \\
	\And
	{Hengyuan Miao} \\
	Tsinghua-Berkeley Shenzhen Institute\\
	Tsinghua Shenzhen International Graduate School\\
 Graduate School, Tsinghua University \\
	Shenzhen, 518055, China \\
 \texttt{miaohy20@mails.tsinghua.edu.cn} \\
	\And
	{Shuang Li} \\
	School of Data Science \\
	The Chinese University of Hong Kong, Shenzhen\\
	Shenzhen 518172, China \\
	\texttt{lishuang@cuhk.edu.cn} \\
}

\hypersetup{
pdftitle={A template for the arxiv style},
pdfsubject={q-bio.NC, q-bio.QM},
pdfauthor={Chao Yang, Hengyuan Miao, Shuang Li},
pdfkeywords={First keyword, Second keyword, More},
}

\begin{document}
\maketitle

\begin{abstract}
We aim to explicitly model the {\it delayed} Granger causal effects based on multivariate Hawkes processes. The idea is inspired by the fact that a causal event usually takes some time to exert an effect. Studying this {\it time lag} itself is of interest. Given the proposed model, we first prove the {\it identifiability} of the delay parameter under mild conditions. 
We further investigate a model estimation method under a complex setting, where we want to {\it infer the posterior distribution of the time lags} and understand how this distribution varies across different scenarios. We treat the time lags as latent variables and formulate a Variational Auto-Encoder (VAE) algorithm to approximate the posterior distribution of the time lags. By explicitly modeling the time lags in Hawkes processes, we add flexibility to the model. The inferred time-lag posterior distributions are of scientific meaning and help trace the original causal time that supports the root cause analysis. We empirically evaluate our model's event prediction and time-lag inference accuracy on synthetic and real data, achieving promising results.
\end{abstract}

\section{Introduction}
\label{introduction}
\setlength{\abovedisplayskip}{0pt}
\setlength{\abovedisplayshortskip}{0pt}
\setlength{\belowdisplayskip}{0 pt}
\setlength{\belowdisplayshortskip}{0 pt}
\setlength{\jot}{0pt}
\setlength{\floatsep}{0ex}
\setlength{\textfloatsep}{0ex}
\setlength{\intextsep}{0ex}
\setlength{\topsep}{0ex}
\setlength{\partopsep}{0ex}
\setlength{\parskip}{0.68ex}

Complex systems often produce voluminous event data with {\it stochastic} and {\it irregularly-spaced} occurrence times. Temporal point processes (TPPs) provide an elegant tool for modeling the dynamics of these event sequences in {\it continuous time}, which directly treat the inter-event time as random variables~\cite{daley2003introduction}. TPPs are characterized by the intensity functions. Among various TPPs, Hawkes processes are a classic and transparent model, with intensity functions being nonnegative functions to capture the {\it triggering effects} from previous events. The intensity functions model the self-exciting and mutual-exciting triggering effects across dimensions subject to an underlying causal graph. There is a surging interest in understanding the Granger causality, i.e., which event triggers which event, and estimate the causal effect based on the multivariate Hawkes processes~\cite{eichler2017graphical,gao2021causal}. 

Although one can design flexible {\it excitatory} TPPs with various triggering kernels, such as nonparametric~\cite{eichler2017graphical} or Gaussian mixture kernels~\cite{xu2016learning}, none of the existing models can explicitly capture the {\it time-lagged} Granger causality. They usually assume that when a causal event happens, its impact on the outcome event's intensity functions is instantaneous. In some cases, {\it it takes time for a cause to exert an effect} and estimating the time lags is essential. For example, during the coronavirus disease pandemic, understanding the {\it incubation periods} for different Omicron variants is a central topic -- when a person first gets exposed to the virus, it usually takes several days before he/her becomes contagious, develops symptoms and tests positive~\cite{quesada2021incubation}. It is of great interest to learn the latent distributions of incubation periods (modeled as time lags) from data so that accurate event prediction, reliable root cause detection and smart intervention can be performed. 

For another example, to curb the spread of covid and reduce the death rate, policy-makers usually conduct a series of actions at different decision times. The government may choose to lock down a city, suspend flights, require facial mask-wearing, encourage the population to be vaccinated, and so on. Some of these interventions may play a more rapid effect, while others may take much longer before they can exert some influence. Learning the time lag distributions of each treatment will provide policy-makers with more informative information and aid future policy optimization. 
\begin{figure}[t]
\centering 
\includegraphics[width=0.95\textwidth]{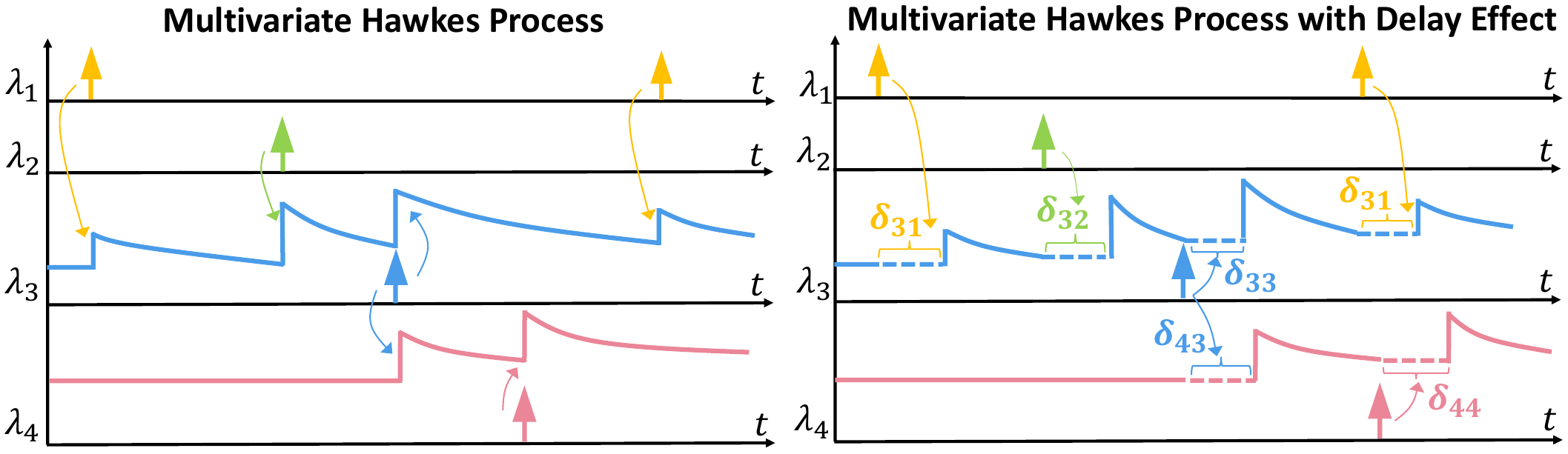} 

\caption{{\small Illustration of the multivariate Hawkes Process without (left) and with (right) delay effects. We assume events 1 and 2 have  triggering effects to event 3, and event 3 has a triggering effect to event 4. Events 3 and 4 both have triggering effects to themselves. The delay effects are denoted as $\delta_{31}, \delta_{32}, \delta_{43}, \delta_{33}, \delta_{44}$ respectively. In the figures, we are considering exponential temporal kernels.}}

\label{fig:multi_hawkes_delayed}
\end{figure}

In this paper, we explicitly introduce the time lag variables to the multivariate Hawkes models, as illustrated in Fig.~\ref{fig:multi_hawkes_delayed}. Given the proposed model, we first prove the {\it identifiability} of the {\it delay parameters} under mild conditions. The widely used exponential kernel $g(t) =a \exp(-\beta(t-\delta))$, with $a$ being the triggering magnitude, $\beta$ being the decaying rate, and $\delta$ being the delay, satisfy the identifiability conditions. We also prove that given the existence of the delays, the $a$ and $\beta$ parameters are also identifiable. It is noteworthy that this identifiability result doesn't depend on specific model estimation methods. One can either use maximum likelihood estimation \cite{ogata1978asymptotic} or moment matching methods \cite{achab2017uncovering}, which will not harm the identifiability results.

In this paper, we especially consider a complex setting, where we treat the {\it time lags} as {\it latent variables} and aim to infer their {\it posterior distributions} given the observational event times. The exact inference is almost intractable. We resort to a VAE learning framework~\cite {kingma2013auto}, where we introduce an amortized neural network function to approximate the time lag posteriors using the observational events as inputs. The posterior inference network and other model parameters are jointly optimized by maximizing the ELBO, which is a lower bound of the marginal likelihood. We will perform the time lag inference and other model parameter estimation simultaneously. We empirically evaluated our estimation method on a synthetic dataset and two healthcare-related datasets. The results demonstrated that when the input event sequences are sufficient, our proposed method  achieve promising results in terms of latent time lag inference, model parameter estimation, and future event prediction. 

Our contributions can be summarized as follows: ({\it i}) We explicitly model the delayed Granger causality for the multivariate Hawkes process and prove the identifiability of the model parameters under mild conditions; ({\it ii}) We propose a customized VAE learning framework that can jointly perform the time lag inference and model parameter estimation. Our related work can be found in Appendix. 

\section{Preliminaries}
\paragraph{Hawkes Processes} Consider a $U$-dimensional multivariate Hawkes process $\{N_u(t)\}_{u=1, \dots, U}$, with intensity functions $\{\lambda_u(t)\}_{u=1, \dots, U}$. Define the union of all the dimensional events up to $t$ as $$\mathcal{H}_t= \left\{t_k^u: 1 \leq k \leq N_u(t), 1 \leq u \leq U\right\}.$$ 
The {\it conditional intensity function} for any dimension $u$ is modeled as
\begin{align}
\lambda_u(t\mid \Hcal_t) = \mu_u + \sum_{u'=1}^U \int_0^{t-} g_{uu'}(t-s) dN_{u'}(s),
\label{eq:Hawkes}
\end{align}
where $\mu_u$ is the base term, and $g_{uu'}(t)\geq 0$ whenever $t\geq 0$ is called impact or triggering function from dimension $u'$ to $u$~\citep{eichler2017graphical}. We can further decompose the triggering function as $g_{uu'}(t) = a_{uu'} \kappa_{uu'}(t)$, $a_{uu'}\geq 0$, to add the interpretability. The support of $a_{uu'}$ indicates the existence of the Granger causal effects from $u'$ to $u$, and the magnitude of $a_{uu'}$ reflects the strength of the effect. 
The temporal kernel function $\kappa_{uu'}(t)$ models how the triggering effect from the events of $u'$ to $u$ will evolve over time. For example, if we use the exponential temporal kernel function, we model that the triggering effect will decay over time as $\kappa_{uu'}(t) =\exp(-\beta_{uu'} t) \mathbbm{1}(t\geq 0)$
where $\beta_{uu'}$ indicates the decaying rate. One can also use  flexible models such as nonparametric~\citep{eichler2017graphical} or neural-based models~\citep{zhang2020cause} to parameterize the intensity function. In this paper, we will focus on the exponential kernel to illustrate the ideas. 

Given the definition of the multivariate Hawkes processes, \cite{eichler2017graphical,xu2016learning} rigorously defined the Granger Causality. Consider a Granger causality graph $G=(\mathcal{U}, \mathcal{E})$, where $\mathcal{U}$ represents the node (i.e., event type) set and $\mathcal{E}$ defines the associated edge set, with the directed edges indicating the causation. We say $u' \rightarrow u \in \mathcal{E}$ if event type $u'$  Granger-causes event type $u$. Specially, we have:
\begin{theorem}[Granger Causality for Multivariate Hawkes Process]\citep{eichler2017graphical,xu2016learning}. Given a Hawkes process defined in Eq.~(\ref{eq:Hawkes}) and a Granger causality graph $G(\mathcal{U}, \mathcal{E})$, if the condition $d N_{u'}(t-s)>0$ for $0 \leq s<t \leq T$ holds, then, $u' \rightarrow u \notin \mathcal{E}$ if and only if $g_{uu'}(t)=0$ for $t \in[0, \infty]$.
\end{theorem}
From the above theorem, the detection of the Granger causality for multivariate Hawkes processes boils down to checking whether $\phi_{uu'}(t)$ is a zero function everywhere. 

\section{Model: Hawkes Processes with Time Lags}
\label{sec:model}
Starting from the basic definition in Eq.~(\ref{eq:Hawkes}), we introduce the {\it time lag parameters} to capture the delayed Granger causality. Define $\bm{\delta} = [\delta_{uu'}]$, where $\delta_{uu'}$ indicates the time-lagged causal effect from event type $u'$ to $u$. The {\it conditional intensity function} for the multivariate (linear) Hawkes process with {\it delayed effect} is defined as
\begin{align}
\lambda^{\text{delay}}_u(t\,|\,\Hcal_t) = \mu_u + \sum_{u'=1}^U \int_0^{t-\delta_{uu'}} g_{uu'}(t-\delta_{uu'}-s) dN_{u'}(s),
\label{eq:Hawkes_delay}
\end{align}
where $g_{uu'}(t)\geq 0$ whenever $t\geq 0$. Let time horizon $t \geq 0$ and assume that event times $\mathcal{H}_t$ are observed in the interval $(0, t]$. Given a parametric model $\theta \in \Theta$, the log-likelihood function $\ell(\theta):= \log p_{\theta}(\mathcal{H}_t)$ is computed as~\cite{daley2003introduction}
\begin{align}
\mathcal{\ell}(\theta) = \sum_{u=1}^U \mathcal{\ell}_u(\theta), \quad \text{with} \quad \mathcal{\ell}_u(\theta)= \sum_{k=1}^{N_u(t)} \log\lambda^{\text{delay}}_{u}(t_k^u \mid\Hcal_{t^u_k}; \theta) - \int_0^t \lambda^{\text{delay}}_{u} (\tau \mid \Hcal_{\tau}; \theta) d\tau .
\label{eq:likelihood}
\end{align}

\section{Identifiability}
\newcommand{\bdelta}{\boldsymbol{\delta}}
\newcommand{\tdelta}{\tilde{\delta}}
\newcommand{\tbdelta}{\tilde{\boldsymbol{\delta}}}
\newcommand{\blambda}{\boldsymbol{\lambda}}
\newcommand{\bmu}{\boldsymbol{\mu}}
\newcommand{\tmu}{\tilde{\mu}}
\newcommand{\tbmu}{\tilde{\boldsymbol{\mu}}}
\newcommand{\balpha}{\boldsymbol{\alpha}}
\newcommand{\talpha}{\tilde{\alpha}}
\newcommand{\tbalpha}{\tilde{\boldsymbol{\alpha}}}
\newcommand{\btheta}{\boldsymbol{\theta}}
\newcommand{\tbtheta}{\tilde{\boldsymbol{\theta}}}
\newcommand{\cH}{\mathcal{H}}
\newcommand{\cD}{\mathcal{D}}
\newcommand{\cU}{\mathcal{U}}
\newcommand{\cK}{\mathcal{K}}
\newcommand{\N}{\mathbb{N}}
\newcommand{\tcU}{\tilde{\mathcal{U}}}
\newcommand{\tcK}{\tilde{\mathcal{K}}}
\newcommand{\tu}{\tilde{u}}

In this section, we discuss the parameter identifiability of the time-lagged Hawkes process defined in Eq.~(\ref{eq:Hawkes_delay}). 
Let $\theta=\left(\theta_1, \ldots, \theta_U\right)$ be the model parameters with 
\begin{align}
\theta_u =\left( \mu_u, \,[\delta_{uu'}]_{u' = 1, \dots, U} \right) \in \Theta,
\end{align}
where $\Theta \subset \mathbb{R}_{+} \times \mathbb{R}_{+}^U$. 
We will also discuss a specific case with the exponential kennel $g(t) =\alpha \exp(-\beta(t-\delta))$, in which case we have
\begin{align}
\theta_u =\left( \mu_u,\, [a_{uu'}]_{u' = 1, \dots, U},\,  [\beta_{uu'}]_{u' = 1, \dots, U}, \,[\delta_{uu'}]_{u' = 1, \dots, U} \right) \in \Theta,
\end{align}
where $\Theta \subset \mathbb{R}_{+} \times \mathbb{R}_{+}^U \times \mathbb{R}_{+}^U \times \mathbb{R}_{+}^U $. 
Define $\lambda^{\text{delay}}(t\mid \mathcal{H}_t; \theta) = \{\lambda^{\text{delay}}_u(t \mid \mathcal{H}_t; \theta_u)\}_{u=1, \dots, U}$. We first provide the conditions that make the delayed Hawkes processes asymptotically stationary, which are the basic conditions for the parameter identifiability. 
\begin{proposition}[Stationarity]
The multivariate delayed (linear) Hawkes process has asymptotically stationary increments and $\lambda^{\text{delay}}(t\mid \mathcal{H}_t)$ is asymptotically stationary if the kernel satisfies the following stability assumption. The matrix 
$$G:=\left[ \left\|g_{uu'}\right\|\right]_{u, u' \in [U]}$$
has a spectral radius smaller than 1. Here, $\left\|g_{uu'}\right\|$ stands for $\left\|g_{uu'}\right\| = \int |g_{uu'}(t)| dt $.
\end{proposition}
For the exponential kernel, the stationary conditions for the delayed linear Hawkes process do not depend on the time lag parameters. Indeed, one can simply compute $\int_0^{+\infty}g_{uu'}(t - \delta_{uu'}) = \int_0^{\delta_{uu'}} 0 dt + \int_{\delta_{uu'} }^{+\infty}a_{uu'} \exp\left( -\beta( t-\delta_{uu'})\right)dt = \frac{a_{uu'}}{\beta}$, which does not depend on $\delta_{uu'}$. In other words, the stationary conditions for the delayed linear Hawkes process is the same with the conditions when there are no delays. 

Next, we discuss identifiability.
\begin{theorem}[Identifiability]\label{thm:id}
Let $\mathcal{H}_t$ be a realization of the multivariate delayed Hawkes process (Eq.~\ref{eq:Hawkes_delay}). 
Assume for every $u,u'\in[U]$, $g_{uu'}$ is positive at 0 and strictly decreasing at a neighborhood of 0, and the inter-arrival time is continuously distributed. 
Then $\lambda^{\text{delay}}(t\mid \mathcal{H}_t; \theta) = \lambda^{\text{delay}}(t\mid \mathcal{H}_t; \tilde{\theta})$ for all $t$ implies that, with probability one, $\mu_u=\tilde{\mu}_u, \delta_{uu'}=\tilde{\delta}_{uu'}$ for all $u,u'=1,\ldots,U$.
In addition, for the exponential kernel, assume $\beta_{uu'}=\beta_u$ for all $u,u'=1,\ldots,U$, and that for every $(u, u')\in\{1,\ldots,U\}^2$, $u \neq u'$, there exists a time interval $[\tau, \tau^+)$, with $\tau$ from the process $N_{u{'}}(t-\delta_{uu'})$ and $\tau^+$ from the process $N_{u{}}(t-\delta_{uu})$, such that it contains arrivals from only the process $N_{u{'}}(t-\delta_{uu'})$. 
Then $\lambda^{\text{delay}}(t\mid \mathcal{H}_t; \theta) = \lambda^{\text{delay}}(t\mid \mathcal{H}_t; \tilde{\theta})$ for all $t$ implies that, with probability one, $a_{uu'}=\tilde{a}_{uu'}$, $\beta_{uu'}=\tilde{\beta}_{uu'}$, $u,u'=1,\ldots,U$.
\end{theorem}

\textbf{Proof Sketch.}
We prove the identifiability of $\mu_u$, $\delta_{uu'}$, $\beta_{u}$ and $a_{uu'}$ in order.
The identifiability of $\mu_u$ holds because of the constant intensity prior to the first arrival and holds without any assumptions stated in the theorem.
The identifiability of $\delta_{uu'}$ is due to the observation that the gap between each pair of arrivals is all different from each other with probability one under our assumptions, and thereby the delay parameters can be uniquely determined by the gaps between jumps of the conditional intensity function.
For the exponential kernel, the identifiability of $\beta_u$ and $a_{uu'}$ can be obtained in a similar way to the argument made in \cite{bonnet2022inference}, in which the assumption facilitates the disentanglement of different arrival processes. 
\endproof

Theorem 3 shows the model parameters for the delayed Hawkes process are identifiable under mild conditions. Note that this identifiability result holds regardless of the model estimation methods. One can freely choose maximum likelihood estimation \cite{ogata1978asymptotic} or moment matching methods \cite{achab2017uncovering}, which will not affect the identifiability conclusion. 

\section{Learning and Inference: Variational Auto-Encoder (VAE)}

In this section, we propose a customized VAE {\it learning} and {\it inference} framework~\cite{kingma2013auto} for the proposed Hawkes process with time lags (Eq.~\ref{eq:Hawkes_delay}). Given the observational events, we treat the {\it time lags} as {\it latent variables} and are interested in {\it inferring} their {\it posterior} distributions. 

\begin{figure*}[ht]
\centering 
\includegraphics[width=0.8\textwidth]{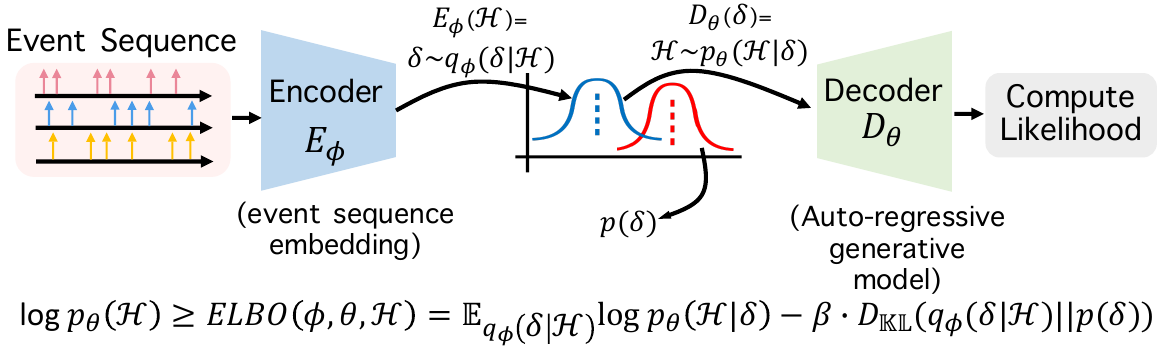} 
\vspace{-10pt}
\caption{{\small VAE learning and inference framework for the Hawkes Process with delayed effects.}}
\label{fig:framework}
\end{figure*}

Specifically, we maximize the (variational) lower bound of the marginal likelihood (Eq.~\ref{eq:likelihood}), known as ELBO,
\begin{align}
\quad \log p_{\theta}(\Hcal_t) \geq \underbrace{-D_{KL} (q_{\phi}(\delta|\Hcal_t) \,||\,p_{\theta}(\delta)) + \EE_{q_{\phi}(\delta|\Hcal_t)}[\log p_{\theta}(\Hcal_t|\bm{\delta})]}_{\text{ELBO}}
\label{eq:ELBO}
\end{align}
where we will use an {\it amortized} recognition neural network $q_{\phi}(\delta|\Hcal_t)$ to approximate the true posterior distributions of the time lags. It's well-known that ELBO becomes tight when the approximated posterior $q_{\phi}(\delta|\Hcal_t)$ matches the true posterior $p_{\theta}(\delta|\mathcal{H}_t)$. In practice, we can introduce a parameter $\beta$ coefficient to the KL term and use the $\beta$-VAE objective ~\citep{higgins2017beta} as shown in Fig.~\ref{fig:framework} to better control the encoding representation capacity of the inference network. 
The following advantages are brought about by our VAE framework: 
\paragraph{({\it i})} This {\it flexible} inference and learning framework for the delayed Hawkes process can capture how the time lag posterior $q_{\phi^*}(\delta|\Hcal_t)$ {\it changes over time or varies across different settings} by {\it conditioning} on observational events. The inference results are of scientific meaning and may help trace the original causal time that supports the root cause analysis. 
\paragraph{({\it ii})} Both the time lag inference parameters and other model parameters will be jointly learned. We will parameterize $p_{\theta}(\delta)$ and $q_{\phi}(\delta|\Hcal_t)$ as some  parametric families so that the first KL divergence in ELBO can be {\it analytically} computed. For the second term in ELBO, we aim to use the {\it reparameterization trick} to have an estimate of the gradient with less variance. All the parameters will be optimized end-to-end through stochastic gradient descent.
\paragraph{Encoder: Event Sequence Embedding} We will use Transformer type of architecture~\citep{vaswani2017attention} to build the probabilistic inference network $q_{\phi}(\delta|\Hcal_t)$, which encodes the input {\it event sequences} to the {\it latent time lag space}. The event sequence embedding is obtained by event time and type embedding similar to Transformer. 
\paragraph{{\it \underline{Event time embedding:}}} Define $\{\bm{z}^{u}_k \}_{k = 1, \dots, N_u(t)}$ as the time embedding of event sequences of dimension $u$. We use the sine and cosine functions of different frequencies $\bar{\bm{z}}_{(t_k^u, 2i)}^u = sin(t_k^u/10000^{2i/d_{model}})$ and $\bar{\bm{z}}_{(t_k^u, 2i+1)}^u = cos(t_k^u/10000^{2i/d_{model}})$ to get the time embedding. Here, $d_{model}$ is the embedding dimension, $t_k^u$ is the $k$-th event time for the event sequence of dimension $u$ and $i$ is the dimension for embedding. 
\paragraph{{\it \underline{Event type embedding:}}} We learn an event type embedding matrix $\bm{U}\in \Rcal^{d_{model}\times U}$, where the $u$-th column of $\bm{U}$ is a $d_{model}$-dimensional embedding for event type $u$.
\paragraph{{\it \underline{Event sequence embedding:}}} We concatenate event time and type embedding to obtain the event embedding. Denote $\{\bm{\tau}^{u}_k \}_{k = 1, \dots, N_u(t)}$ as the sequence of event embedding of dimension $u$ and the concatenated matrix as $\bm{\tau}^{(u)}\in R^{2d_{model}\times N_u(t)}$. We will assume the time lags are Gaussian- or exponential-distributed and map the sequence embedding to the corresponding parameter space of $\delta$. For example, we can let 
$ \gamma = \phi(\bm{\tau}^{(u) \top}\bm{\tau}^{(u')})$, where $\phi$ is a nonlinear function with learnable parameters and $\gamma$ is the rate of the posterior exponential distribution for $\delta_{uu'}\mid \mathcal{H}_t$.
\paragraph{Decoder: Auto-regressive Model} Now, we can sample $\delta=[\delta_{uu'}]$, $\delta_{uu'}\geq 0$ from the latent time lag space, and build our decoder, which is our proposed delayed multivariate Hawkes process, $p_{\theta}(\Hcal_t|\delta)$. The intensity function can be evaluated in an auto-regressive way by conditioning on the sampled delays and observational events as shown in Eq.~(\ref{eq:Hawkes_delay}). Given the conditional intensity, the log-likelihood can be evaluated by Eq.~(\ref{eq:likelihood}) and we can approximate $\EE_{q_{\phi}(\delta|\Hcal_t)}[\log p_{\theta}(\Hcal_t|\delta)]$ by drawing samples from the latent time lag space. 
\paragraph{Stochastic Gradient Variational Bayes}
We aim to learn all the parameters, including encoder approximate posterior model parameters and the decoder multivariate Hawkes model parameters end-to-end in a differentiable way, by maximizing the ELBO (Eq.~(\ref{eq:ELBO})). Considering $\delta\geq 0$, we may parameterize both $p_{\theta}(\delta)$ and $q_{\phi}(\delta|\Hcal_t)$ as exponential distributions so that the first KL term of ELBO can be analytically evaluated. For the second term of ELBO, we will use the reparameterization trick for the sampling of time lags $\delta$ to make the estimate of the gradient less noisy. For example, suppose we use the exponential distribution of the latent time lags, we can compute the sample as $ \tilde{\delta}_{uu_j} = \frac{1}{ \gamma}\cdot \epsilon$, with $\epsilon\sim \exp(1)$ and $\gamma = \phi(\bm{\tau}^{(u) \top}\bm{\tau}^{(u')})$. We can then form the Monte Carlo estimate of the expectation $\EE_{q_{\phi}(\delta|\Hcal_t)}[\log p_{\theta}(\Hcal_t|\delta)]$ as a deterministic function of $\phi$ by changing the expectation w.r.t. $p(\epsilon)$. Now, we can optimize the ELBO objective using stochastic gradient descents, where the inference parameters $\phi$ and the model parameters $\theta$ will be jointly updated. 
\paragraph{Event Prediction} Another nice property of this VAE framework is that the decoder is a generative model, with which we can generate future (fake) events as predictions. For example, given the historical event sequences $\Hcal_{t_n}$ at hand, we can first feed $\Hcal_{t_n}$ to the encoder to get a time lag posterior, $q_{\hat{\phi}}(\delta|\mathcal{H}_{t_n}) $, and then sample time lags from this posterior to generate future events using $p_{\hat{\theta}}(\hat{t}_{n+1}, \hat{u}_{n+1}|\delta, \mathcal{H}_{t_n})$. The generative process can be executed in an auto-regressive way. We can repeat this procedure multiple times to get a more accurate prediction. 
\section{Experiments}
To validate the inference and learning accuracy of our method, we first evaluate our method on synthetic datasets with {\it known} time lags and other parameters. We then consider two {\it healthcare} datasets to evaluate event prediction accuracy of our method and compare with other baselines. 
\subsection{Synthetic Data Experiments}
\subsubsection{Experiment Setup}
For the delayed Hawkes process, an exponential kernel $g(t) =a \exp(-\beta(t-\delta))$ is used to build the intensity functions. We assume a  {\it Gaussian prior} and {\it posterior} distributions for the latent time lags. To better show the model performance, we assume the sparsity structure of the {\it Granger causal graph} (i.e., no granger causality for some dimension pairs) is known as priori. 2000 samples of event sequences were used as training data and each sequence has 10 dimensions. One can imagine that given more samples the {\it learning} and {\it inference} performance will improve as a result. 

\subsubsection{Results}

\begin{figure}[ht]
\centering 
\includegraphics[width=1\textwidth]{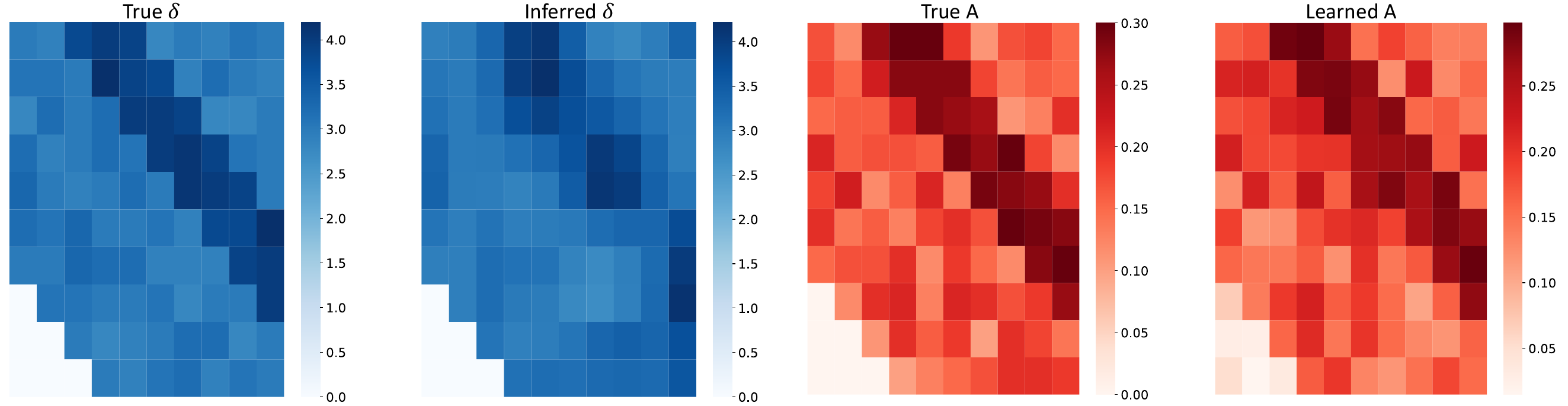} 
\caption{{\small (Red) True and learned A matrix. (Blue) True and inferred time lags: mean parameter of Gaussian distribution. The sparsity pattern (no Granger causality for the dimension pairs in the lower left corner) is unknown when learning A matrix but given as prior knowledge when inferring delayed causal effect.}}
\label{fig:VAE_normal_pattern}
\end{figure}

\paragraph{Inference} We compared the {\it inferred time lags} (posterior mean) with the true time lags in Fig.~\ref{fig:VAE_normal_pattern}. From the results, we conclude that our method achieves satisfactory performance in inferring the latent time lags. The patterns of larger delayed causal effects are clearly inferred. For dimension pairs with no delayed causal effect (lower left corner), the inferred delayed causal effect was closed to zero. Note that these figures only show the mean parameter of the Gaussian distribution, although we can infer the entire posterior distribution.

\paragraph{Learning} We also compared the {\it model parameter learning} results with the ground truth. For multivariate Hawkes processes, the model parameters include base term $\mu$, impact parameter $a$, and decay parameter $\beta$ for the exponential temporal kernel. Here we only show the learned impact parameters $A=[a_{uu'}]$ for a dense causal graph example, as shown in Fig.~\ref{fig:VAE_normal_pattern}. Our model successfully uncovered the impact parameter matrix with most patterns clearly uncovered.
\paragraph{Ablation Study} We consider four ablation studies: (a) use VAE but remove the event type embedding, (b) use VAE but remove both event time embedding and event type embedding, (c) use variational inference, which directly treats the parameters of time lag distribution as learnable parameters, and (d) take the parameters of time lag distribution as learnable parameters and learn these parameters by optimizing the likelihood. The metric we use is the "absolute error rate", which is defined as $\varepsilon = \frac{1}{N}\sum_{n=1}^N \frac{\left | L_n - T_n \right |}{T_n} \times 100\%$, where $T$ denotes the ground truth value of parameters, $L$ denotes the learned value of parameters, and $N$ denotes the total number of dimension pairs of a specific model parameter. A lower absolute error rate indicates better model efficacy. As indicated by Tab.~\ref{tab:ablation_study}, removing the above-mentioned modules or simplifying the optimizing process deteriorate the model performance. Hence, these modules are critical to guarantee the proposed model's efficacy.


\begin{table*}[t]
\centering
 \begin{tabular}{c|cccc}
\hline
\diagbox{Model}{Params} & base: $\mu$ & impact: $A$ & decay: $\beta$ & delay: $\delta$ \\
\hline
Model $(a)$ & 10.12\% & 23.78\% & \textbf{40.25\%} & 9.78\% \\
\hline
Model $(b)$ & 18.57\% & 35.76\% & 64.69\% & 53.58\% \\
\hline
Model $(c)$ & 12.35\% & 25.43\% & 48.65\% & 13.28\% \\
\hline
Model $(d)$ & 14.58\% & 27.54\% & 56.55\% & 26.88\% \\
\hline
\textbf{OUR METHOD} & \textbf{9.21\%} & \textbf{18.39\%} & 43.65\% & \textbf{5.63\%} \\
\hline
\end{tabular}
\caption{{\small Summary of the ablation study results. When computing the absolute error rate, we ignore the dimension pairs with no delay causal effect to prevent numerical issues. For $\delta$, we only compute the absolute error rate for the mean parameter of Gaussian distribution.}}
\label{tab:ablation_study}
\end{table*}

\paragraph{Event Prediction} We also predict the event time for two known event types (indicated by dim-1 and dim-2) and use the root mean squared error (RMSE) metric to evaluate. Lower RMSE (unit is hour in this example) indicates better performance. In Appendix, we introduced the baselines used in this paper. As shown in Tab.~\ref{tab:prediction_event_time_synthetic_mimic_and_covid_data}, for dim-2, our model outperforms all the baselines in this experiment. For dim-1, our model also performs well and the RMSE is the third lowest among all baselines.

\begin{table*}[t]
\centering
 \begin{tabular}{c|c|c|c|c}
\toprule
Dataset & \multicolumn{2}{c|}{Synthetic Data} & \multicolumn{2}{c}{Real-World Data} \\


\midrule
\midrule

Method & Dim-1 & Dim-2 & Normal Urine & Dropping Infection \\
\hline

GCH & 3.1585 & 3.3474 & 3.4686 & 12.2378   \\
\hline

LG-NPP & 2.8623 & 3.1345 & 3.7272 & 12.5685  \\
\hline

GM-NLF & 3.1566 & 3.2389 & 4.3216 & 13.5734  \\
\hline

THP & \bf{2.1078} & 2.2248 & 3.1794 & 11.2465 \\
\hline

RMTPP & 2.1575 & 2.3587 & 3.7684 & 12.0162 \\
\hline

ERPP & 2.4357 & 2.5138 & 3.6578 & 12.8767 \\
\hline

MLE & 2.7586 & 3.0155 & 4.1123 & 12.8234  \\
\hline

VI & 2.6523 & 2.7254 & 3.5217 & 10.8255  \\
\hline

\textbf{OUR METHOD} & 2.3634 & \textbf{2.2043} & \bf{2.9324} & \bf{9.8946}  \\

\bottomrule

\end{tabular}
\caption{{\small Event time prediction RMSE on synthetic data, MIMIC-IV data (evaluating prediction of occur time of Normal Urine) and Covid Policy Tracker data (evaluating prediction of occur time of dropping confirmed cases/infections)}}
\label{tab:prediction_event_time_synthetic_mimic_and_covid_data}
\end{table*}
\subsection{Healthcare Data Experiments}
\subsubsection{MIMIC-IV Data}
MIMIC-IV\footnote{\url{https://mimic.mit.edu/}} is an electronic health record dataset of patients admitted to the intensive care unit (ICU)~\citep{johnson2023mimic}. We considered patients diagnosed with sepsis \citep{saria2018individualized}, since sepsis is one of the major causes of mortality in ICU, mainly due to septic shock. Septic shocks are medical emergencies and early recognition and treatment would improve survival.

\paragraph{Treatments} The vasopressor therapy is a fundamental treatment of septic-shock-induced hypotension as it aims at correcting the vascular tone depression and then improving organ perfusion pressure. Antibiotics also should be given within a few hours of the diagnosis of sepsis. Some auxiliary
treatments such as packed red blood cells and invasive ventilation are also necessary in ICU.
Therefore, we consider above-mentioned three kinds of treatments: vasopressors, antibiotics and auxiliary treatments in this experiment. Previous studies suggest that there still exists clinical controversy about the delay effect of the treatment and drug usage for the patients. 

\paragraph{Outcome} We treated real time urine as the outcome indicator since low urine is the direct indicator of bad circulatory systems and the signal for septic shock. In contrast, normal urine reflects the effect of the drugs and treatments and the improvement of the patients' physical condition. Some treatments will have a rapid effect on the urine while others might take longer to exert an effect. We assume that the delay effect of each treatment is Gaussian distributed. 

\begin{figure}[H]
\centering 
\includegraphics[width=1\textwidth]{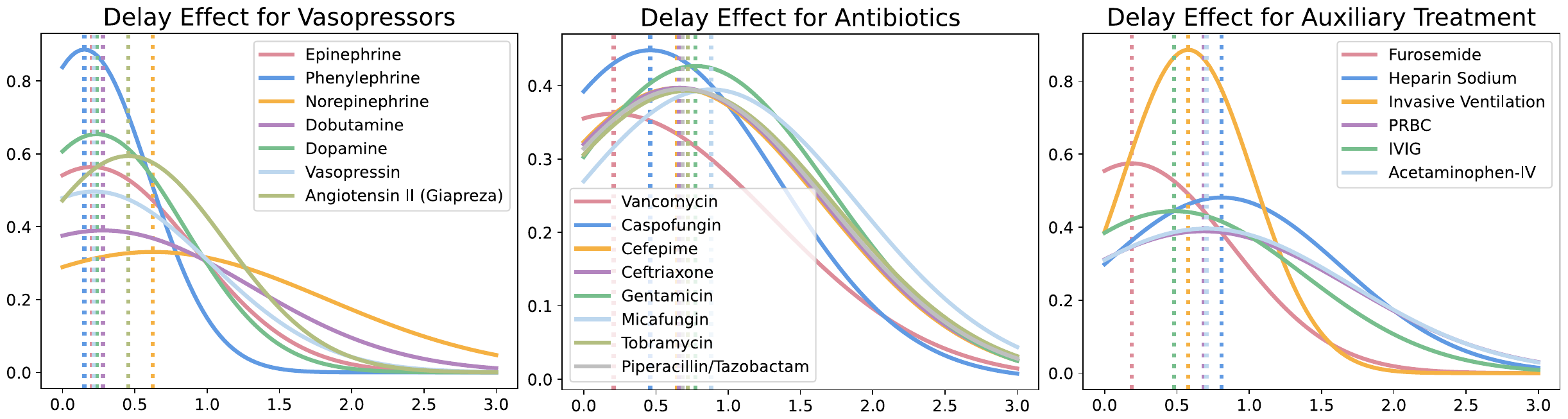}
\caption{\small{Inferred lag posteriors of Vasopressors, Antibiotics and Auxiliary treatments on normal urine output for the population.}}
\label{fig:mimic_delayed_effect}
\end{figure}

\paragraph{Inferred Time Lags} The inferred lag posterior of delay effects for vasopressors are displayed in Fig.~\ref{fig:mimic_delayed_effect} (left). Phenylephrine is a commonly used vasopressor in the neurologic ICU. It works fast with small variance. The delay effect of Epinephrine, Dobutamine, Dopamine, Vasopressin are similar, all around 0.3 hours. Interestingly, although experts recommend using norepinephrine as the first-line vasopressor in septic shock, it has the largest delay effect, which is around 0.6 hours.

As also shown in Fig.~\ref{fig:mimic_delayed_effect} (middle), the delay effect of antibiotics are similar and all around 0.7 hours. This is also consistent with the clinical observation and multiple antibiotics are often used concurrently due to this property. It is worth noting that vancomycin, which is a main therapy for treating sepsis infection caused by Methicillin-resistant Staphylococcus aureus, is the fastest-acting antibiotic, takes effect within 0.3 hours. But its probability density function curve is flat, indicating high variance.

\begin{figure}[H]
\centering 
\includegraphics[width=1\textwidth]{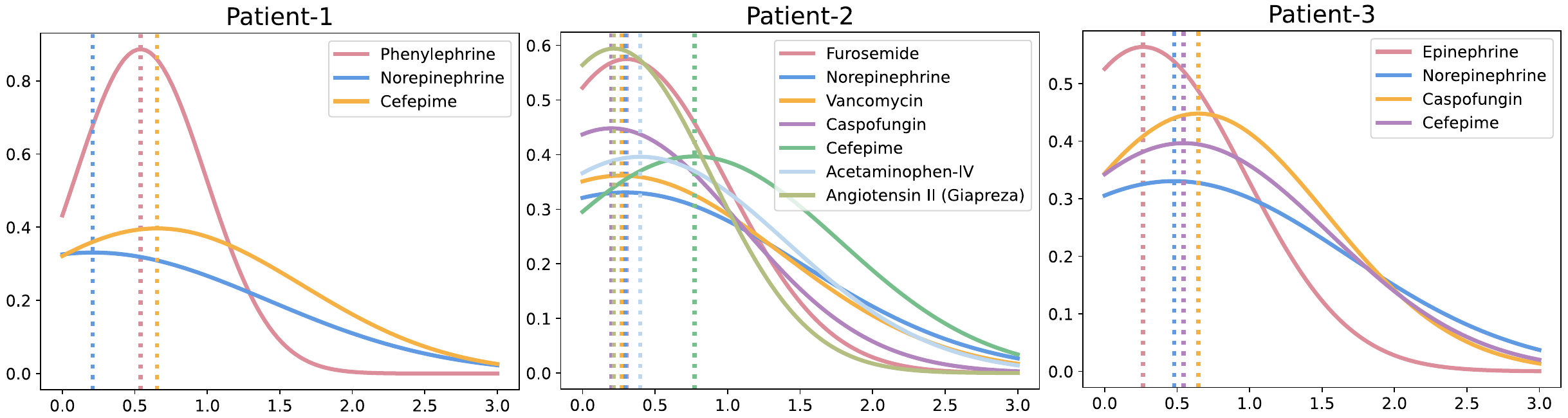}
\caption{\small{Inferred lag posteriors of Vasopressors, Antibiotics and Auxiliary treatments on normal urine output for different patients in ICU. For each patient, only the actual treatment used will be displayed.}}
\label{fig:mimic_delayed_effect_patient}
\end{figure}

For the inferred lag posterior of the delay effect of auxiliary treatments, which are displayed in Fig.~\ref{fig:mimic_delayed_effect} (right), invasive ventilation is a potentially lifesaving intervention for acutely ill patients. It often takes effect within 0.6 hours, and this delay effect is similar across patients, since the variance of the learned Gaussian distribution is small. Similar patterns appear when using Heparin Sodium, Packed Red Blood Cells (PRBC), IV Immune Globulin (IVIG), and Acetaminophen-IV. Furosemide shows clearly short-term response, but the time it takes effect varies widely among different patients.
\paragraph{Patient-Specific Case Study} Our proposed method can provide the inferred lag posterior of delay effects of different treatments for specific patients. The delay effect of treatments for different patients is slightly different but overall similar to the population level. As shown in Fig.~\ref{fig:mimic_delayed_effect_patient}, Norepinephrine produces the desired at the soonest for patient-1. More treatments were attempted for patient-2. Angiotensin II and Caspofungin have the lowest delay effect. The treatment strategy of patient-3 is similar to patient-1, but this patient seems not sensitive to Norepinephrine since it takes a longer time to take effect.
\paragraph{Event Prediction} We compared several SOTA baselines in terms of event prediction. RMSE was used as the metric for evaluating the prediction of occur time of \emph{Normal Urine} event. The performance of our model and all baselines are compared in Tab.\ref{tab:prediction_event_time_synthetic_mimic_and_covid_data}, from which one can observe that our model outperforms all the baselines in this
experiment.

\subsubsection{Covid Policy Tracker}
Covid-19 is an unprecedented pandemic that transformed the world and shaped society with its destructive health and economic impact. Various prevention and control measures have been introduced to curb the spread of the virus. Governments are taking a wide range of measures in response to the COVID-19 outbreak. The Oxford Covid-19 Government Response Tracker (OxCGRT) \footnote{\url{https://github.com/OxCGRT/covid-policy-tracker/}} collects information on which governments have taken which measures, and when ~\citep{hale2021global, hale2020oxford}. This can help decision-makers and citizens understand governmental responses in a consistent way, aiding efforts to fight the pandemic. 

Government policy information in OxCGRT dataset is coded in 25 indicators and organized into 4 categories, which could be referred to Appendix for detailed information. Apart from this, OxCGRT dataset also contains information on confirmed Covid cases in each day. To avoid daily noise, we recorded the cumulative numbers of confirmed cases per 7 consecutive days to represent the trend of the epidemic spread. With the aim of understanding the waiting time for each policy to work, we extracted the time points when the cases started dropping. We selected 7 countries from four continents and focused on the policy data of the year 2021, in which all these 7 countries responded most frequently to outbreaks. Because China had the strictest epidemic prevention policy and has the largest population in the world, we only show the results of China data and that of the other 6 countries could be found in Appendix.

\begin{figure}[H]
\centering 
\includegraphics[width=1\textwidth]{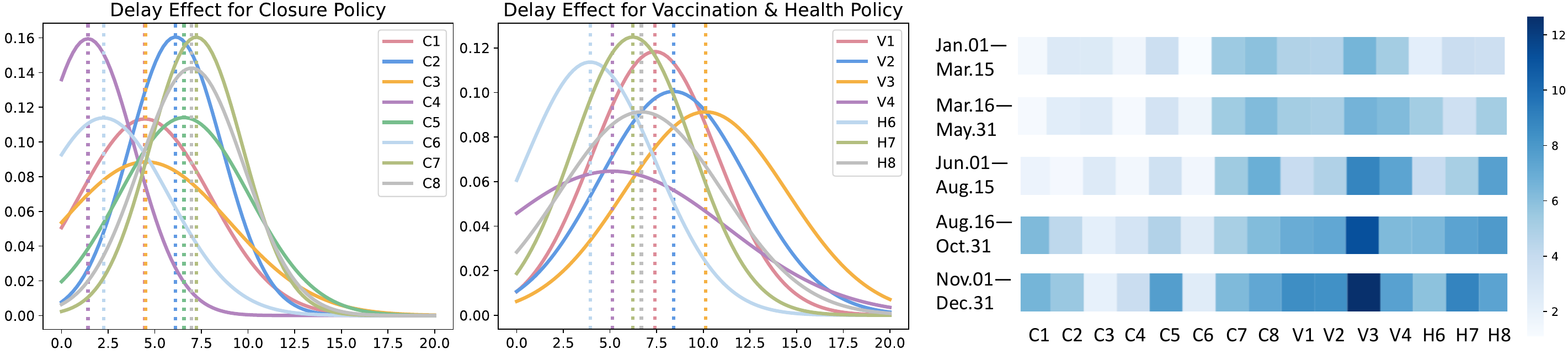} 
\caption{\small{Left two indicate the inferred lag posterior of policy on reduced confirmed cases using Covid-19 Policy Tracker data in China. The right heatmap indicates the change of delay effect over time in the year 2021.}}
\label{fig:covid_delayed_effect_China}
\end{figure}

\paragraph{Inferred Time Lags} In the year 2021, the Chinese government has issued a total of 15 epidemic prevention policies. Closure Policies in China Fig.~\ref{fig:covid_delayed_effect_China} (left) produced the desired results. The reduction of confirmed cases showed an immediate response to Restrictions on gatherings (C4) and Stay at home (C6). About 2 days after these policies are promulgated, the number of new confirmed cases will be sharply reduced. School closing (C1) and Cancel public events (C3) are also appropriate policies, with a delay effect of around 5 days. Surprisingly, despite the large number of labors in China, Workplace closing (C2) did not work as fast as expected. It usually takes a time lag of around 6 days to be effective. 

For Health System Policies in China, Fig.~\ref{fig:covid_delayed_effect_China} (middle), facial coverings (H6) would fleetly help control the infection within only 4 days and with relatively small variance in different provinces. Vaccination policy (H7) and protection of elderly people (H8) also showed similar patterns. The Vaccination Policies in China, Fig.~\ref{fig:covid_delayed_effect_China} (middle), both mandatory vaccination (V4) and vaccine prioritization (V1) exhibited some quick results, which would take effect within 6 and 8 days respectively, indicating that reasonable and orderly vaccination should be an effective way to control the pandemic of Covid-19 in China. However, vaccine financial support (V3) does not promise quick results.
\paragraph{Change of Delay Effect over Time} The dropping of confirmed cases might actually be caused by a complex interplay of various factors. In addition, the degree of influence between events must have varied from time to time. We truncated the data of China into 5 parts according to time in the year 2021 to analyze these potential phenomena. Seeing from Fig.~\ref{fig:covid_delayed_effect_China} (right), the inferred time lag of Closure Policies is almost consistent in 2021. However, the delay effects of Health System Policies and Vaccination Policies are small in the first half of the year 2021 but large in the second half of the year 2021.
\paragraph{Event Prediction} Same SOTA baselines were compared and RMSE metric was also used in this experiment as shown in Tab.~\ref{tab:prediction_event_time_synthetic_mimic_and_covid_data}. Our model still outperformed all the baselines.

\section{Limitation}
The inference accuracy of the time lags depends on the choice of the prior distribution. In practice, we need to leverage domain knowledge and be more careful in choosing the prior distribution, especially when the sample size is small. 
\section{Conclusion}
In this paper, we proposed a flexible time-lagged Hawkes process model. Based on our proposed model, we first proved the identifiability of the model parameters under mild conditions. We further design a customized VAE to jointly perform the time lag inference and model learning. The inferred time-lag posterior distributions are of scientific meaning and help trace the original causal time that supports the root cause analysis.




\bibliographystyle{unsrtnat}
\bibliography{references.bib}  

\begin{thebibliography}{39}
\providecommand{\natexlab}[1]{#1}
\providecommand{\url}[1]{\texttt{#1}}
\expandafter\ifx\csname urlstyle\endcsname\relax
  \providecommand{\doi}[1]{doi: #1}\else
  \providecommand{\doi}{doi: \begingroup \urlstyle{rm}\Url}\fi

\bibitem[Daley et~al.(2003)Daley, Vere-Jones, et~al.]{daley2003introduction}
Daryl~J Daley, David Vere-Jones, et~al.
\newblock \emph{An introduction to the theory of point processes: volume I:
  elementary theory and methods}.
\newblock Springer, 2003.

\bibitem[Eichler et~al.(2017)Eichler, Dahlhaus, and
  Dueck]{eichler2017graphical}
Michael Eichler, Rainer Dahlhaus, and Johannes Dueck.
\newblock Graphical modeling for multivariate hawkes processes with
  nonparametric link functions.
\newblock \emph{Journal of Time Series Analysis}, 38\penalty0 (2):\penalty0
  225--242, 2017.

\bibitem[Gao et~al.(2021)Gao, Subramanian, Bhattacharjya, Shou, Mattei, and
  Bennett]{gao2021causal}
Tian Gao, Dharmashankar Subramanian, Debarun Bhattacharjya, Xiao Shou, Nicholas
  Mattei, and Kristin~P Bennett.
\newblock Causal inference for event pairs in multivariate point processes.
\newblock \emph{Advances in Neural Information Processing Systems},
  34:\penalty0 17311--17324, 2021.

\bibitem[Xu et~al.(2016)Xu, Farajtabar, and Zha]{xu2016learning}
Hongteng Xu, Mehrdad Farajtabar, and Hongyuan Zha.
\newblock Learning granger causality for hawkes processes.
\newblock In \emph{International conference on machine learning}, pages
  1717--1726. PMLR, 2016.

\bibitem[Quesada et~al.(2021)Quesada, L{\'o}pez-Pineda, Gil-Guill{\'e}n,
  Arriero-Mar{\'\i}n, Guti{\'e}rrez, and
  Carratala-Munuera]{quesada2021incubation}
JA~Quesada, A~L{\'o}pez-Pineda, VF~Gil-Guill{\'e}n, JM~Arriero-Mar{\'\i}n,
  F~Guti{\'e}rrez, and C~Carratala-Munuera.
\newblock Incubation period of covid-19: A systematic review and meta-analysis.
\newblock \emph{Revista Cl{\'\i}nica Espa{\~n}ola (English Edition)},
  221\penalty0 (2):\penalty0 109--117, 2021.

\bibitem[Ogata et~al.(1978)]{ogata1978asymptotic}
Yosihiko Ogata et~al.
\newblock The asymptotic behaviour of maximum likelihood estimators for
  stationary point processes.
\newblock \emph{Annals of the Institute of Statistical Mathematics},
  30\penalty0 (1):\penalty0 243--261, 1978.

\bibitem[Achab et~al.(2017)Achab, Bacry, Ga{\i}ffas, Mastromatteo, and
  Muzy]{achab2017uncovering}
Massil Achab, Emmanuel Bacry, St{\'e}phane Ga{\i}ffas, Iacopo Mastromatteo, and
  Jean-Fran{\c{c}}ois Muzy.
\newblock Uncovering causality from multivariate hawkes integrated cumulants.
\newblock In \emph{International Conference on Machine Learning}, pages 1--10.
  PMLR, 2017.

\bibitem[Kingma and Welling(2013)]{kingma2013auto}
Diederik~P Kingma and Max Welling.
\newblock Auto-encoding variational bayes.
\newblock \emph{arXiv preprint arXiv:1312.6114}, 2013.

\bibitem[Zhang et~al.(2020{\natexlab{a}})Zhang, Panum, Jha, Chalasani, and
  Page]{zhang2020cause}
Wei Zhang, Thomas Panum, Somesh Jha, Prasad Chalasani, and David Page.
\newblock Cause: Learning granger causality from event sequences using
  attribution methods.
\newblock In \emph{International Conference on Machine Learning}, pages
  11235--11245. PMLR, 2020{\natexlab{a}}.

\bibitem[Bonnet et~al.(2022)Bonnet, Herrera, and Sangnier]{bonnet2022inference}
Anna Bonnet, Miguel~Martinez Herrera, and Maxime Sangnier.
\newblock Inference of multivariate exponential hawkesprocesses with inhibition
  and application toneuronal activity.
\newblock \emph{arXiv preprint arXiv:2205.04107}, 2022.

\bibitem[Higgins et~al.(2017)Higgins, Matthey, Pal, Burgess, Glorot, Botvinick,
  Mohamed, and Lerchner]{higgins2017beta}
Irina Higgins, Loic Matthey, Arka Pal, Christopher Burgess, Xavier Glorot,
  Matthew Botvinick, Shakir Mohamed, and Alexander Lerchner.
\newblock beta-vae: Learning basic visual concepts with a constrained
  variational framework.
\newblock In \emph{International conference on learning representations}, 2017.

\bibitem[Vaswani et~al.(2017)Vaswani, Shazeer, Parmar, Uszkoreit, Jones, Gomez,
  Kaiser, and Polosukhin]{vaswani2017attention}
Ashish Vaswani, Noam Shazeer, Niki Parmar, Jakob Uszkoreit, Llion Jones,
  Aidan~N Gomez, {\L}ukasz Kaiser, and Illia Polosukhin.
\newblock Attention is all you need.
\newblock \emph{Advances in neural information processing systems}, 30, 2017.

\bibitem[Johnson et~al.(2023)Johnson, Bulgarelli, Shen, Gayles, Shammout,
  Horng, Pollard, Moody, Gow, Lehman, et~al.]{johnson2023mimic}
Alistair~EW Johnson, Lucas Bulgarelli, Lu~Shen, Alvin Gayles, Ayad Shammout,
  Steven Horng, Tom~J Pollard, Benjamin Moody, Brian Gow, Li-wei~H Lehman,
  et~al.
\newblock Mimic-iv, a freely accessible electronic health record dataset.
\newblock \emph{Scientific Data}, 10\penalty0 (1):\penalty0 1--9, 2023.

\bibitem[Saria(2018)]{saria2018individualized}
Suchi Saria.
\newblock Individualized sepsis treatment using reinforcement learning.
\newblock \emph{Nature medicine}, 24\penalty0 (11):\penalty0 1641--1642, 2018.

\bibitem[Hale et~al.(2021)Hale, Angrist, Goldszmidt, Kira, Petherick, Phillips,
  Webster, Cameron-Blake, Hallas, Majumdar, et~al.]{hale2021global}
Thomas Hale, Noam Angrist, Rafael Goldszmidt, Beatriz Kira, Anna Petherick,
  Toby Phillips, Samuel Webster, Emily Cameron-Blake, Laura Hallas, Saptarshi
  Majumdar, et~al.
\newblock A global panel database of pandemic policies (oxford covid-19
  government response tracker).
\newblock \emph{Nature human behaviour}, 5\penalty0 (4):\penalty0 529--538,
  2021.

\bibitem[Hale et~al.(2020)Hale, Webster, Petherick, Phillips, and
  Kira]{hale2020oxford}
Thomas Hale, S~Webster, A~Petherick, T~Phillips, and B~Kira.
\newblock Oxford covid-19 government response tracker (oxcgrt).
\newblock \emph{Last updated}, 8:\penalty0 30, 2020.

\bibitem[Granger(1969)]{granger1969investigating}
Clive~WJ Granger.
\newblock Investigating causal relations by econometric models and
  cross-spectral methods.
\newblock \emph{Econometrica: journal of the Econometric Society}, pages
  424--438, 1969.

\bibitem[Dahlhaus and Eichler(2003)]{dahlhaus2003causality}
Rainer Dahlhaus and Michael Eichler.
\newblock Causality and graphical models in time series analysis.
\newblock \emph{Oxford Statistical Science Series}, pages 115--137, 2003.

\bibitem[Didelez(2008)]{didelez2008graphical}
Vanessa Didelez.
\newblock Graphical models for marked point processes based on local
  independence.
\newblock \emph{Journal of the Royal Statistical Society: Series B (Statistical
  Methodology)}, 70\penalty0 (1):\penalty0 245--264, 2008.

\bibitem[Du et~al.(2016)Du, Dai, Trivedi, Upadhyay, Gomez-Rodriguez, and
  Song]{du2016recurrent}
Nan Du, Hanjun Dai, Rakshit Trivedi, Utkarsh Upadhyay, Manuel Gomez-Rodriguez,
  and Le~Song.
\newblock Recurrent marked temporal point processes: Embedding event history to
  vector.
\newblock In \emph{Proceedings of the 22nd ACM SIGKDD international conference
  on knowledge discovery and data mining}, pages 1555--1564, 2016.

\bibitem[Xiao et~al.(2017)Xiao, Yan, Yang, Zha, and Chu]{xiao2017modeling}
Shuai Xiao, Junchi Yan, Xiaokang Yang, Hongyuan Zha, and Stephen Chu.
\newblock Modeling the intensity function of point process via recurrent neural
  networks.
\newblock In \emph{Proceedings of the AAAI Conference on Artificial
  Intelligence}, volume~31, 2017.

\bibitem[Mei and Eisner(2017)]{mei2017neural}
Hongyuan Mei and Jason~M Eisner.
\newblock The neural hawkes process: A neurally self-modulating multivariate
  point process.
\newblock \emph{Advances in neural information processing systems}, 30, 2017.

\bibitem[Xiao et~al.(2019)Xiao, Yan, Farajtabar, Song, Yang, and
  Zha]{xiao2019learning}
Shuai Xiao, Junchi Yan, Mehrdad Farajtabar, Le~Song, Xiaokang Yang, and
  Hongyuan Zha.
\newblock Learning time series associated event sequences with recurrent point
  process networks.
\newblock \emph{IEEE transactions on neural networks and learning systems},
  30\penalty0 (10):\penalty0 3124--3136, 2019.

\bibitem[Gollob and Reichardt(1987)]{gollob1987taking}
Harry~F Gollob and Charles~S Reichardt.
\newblock Taking account of time lags in causal models.
\newblock \emph{Child development}, pages 80--92, 1987.

\bibitem[Ye et~al.(2015)Ye, Deyle, Gilarranz, and
  Sugihara]{ye2015distinguishing}
Hao Ye, Ethan~R Deyle, Luis~J Gilarranz, and George Sugihara.
\newblock Distinguishing time-delayed causal interactions using convergent
  cross mapping.
\newblock \emph{Scientific reports}, 5\penalty0 (1):\penalty0 1--9, 2015.

\bibitem[Amornbunchornvej et~al.(2019)Amornbunchornvej, Zheleva, and
  Berger-Wolf]{amornbunchornvej2019variable}
Chainarong Amornbunchornvej, Elena Zheleva, and Tanya~Y Berger-Wolf.
\newblock Variable-lag granger causality for time series analysis.
\newblock In \emph{2019 IEEE International Conference on Data Science and
  Advanced Analytics (DSAA)}, pages 21--30. IEEE, 2019.

\bibitem[Absar and Zhang(2021)]{absar2021discovering}
Saima Absar and Lu~Zhang.
\newblock Discovering time-invariant causal structure from temporal data.
\newblock In \emph{Proceedings of the 30th ACM International Conference on
  Information \& Knowledge Management}, pages 2807--2811, 2021.

\bibitem[Zhang et~al.(2020{\natexlab{b}})Zhang, Kuang, Peissig, and
  Page]{zhang2020adverse}
Wei Zhang, Zhaobin Kuang, Peggy Peissig, and David Page.
\newblock Adverse drug reaction discovery from electronic health records with
  deep neural networks.
\newblock In \emph{Proceedings of the ACM Conference on Health, Inference, and
  Learning}, pages 30--39, 2020{\natexlab{b}}.

\bibitem[Wei et~al.(2022)Wei, Xie, Josef, and Kamaleswaran]{wei2022granger}
Song Wei, Yao Xie, Christopher~S Josef, and Rishikesan Kamaleswaran.
\newblock Granger causal chain discovery for sepsis-associated derangements via
  multivariate hawkes processes.
\newblock \emph{arXiv preprint arXiv:2209.04480}, 2022.

\bibitem[Dhurandhar(2010)]{dhurandhar2010learning}
Amit Dhurandhar.
\newblock Learning maximum lag for grouped graphical granger models.
\newblock In \emph{2010 IEEE International Conference on Data Mining
  Workshops}, pages 217--224. IEEE, 2010.

\bibitem[Knapp and Carter(1976)]{knapp1976generalized}
Charles Knapp and Glifford Carter.
\newblock The generalized correlation method for estimation of time delay.
\newblock \emph{IEEE transactions on acoustics, speech, and signal processing},
  24\penalty0 (4):\penalty0 320--327, 1976.

\bibitem[Sakurai et~al.(2005)Sakurai, Papadimitriou, and
  Faloutsos]{sakurai2005braid}
Yasushi Sakurai, Spiros Papadimitriou, and Christos Faloutsos.
\newblock Braid: Stream mining through group lag correlations.
\newblock In \emph{Proceedings of the 2005 ACM SIGMOD international conference
  on Management of data}, pages 599--610, 2005.

\bibitem[Shiller(1973)]{shiller1973distributed}
Robert~J Shiller.
\newblock A distributed lag estimator derived from smoothness priors.
\newblock \emph{Econometrica: journal of the Econometric Society}, pages
  775--788, 1973.

\bibitem[Zhou et~al.(2017)Zhou, Hong, Xing, Bian, Xie, and
  Xu]{zhou2017discovering}
Xiabing Zhou, Haikun Hong, Xingxing Xing, Kaigui Bian, Kunqing Xie, and
  Mingliang Xu.
\newblock Discovering spatio-temporal dependencies based on time-lag in
  intelligent transportation data.
\newblock \emph{Neurocomputing}, 259:\penalty0 76--84, 2017.

\bibitem[Du et~al.(2017)Du, Song, Han, and Hong]{du2017temporal}
Sizhen Du, Guojie Song, Lei Han, and Haikun Hong.
\newblock Temporal causal inference with time lag.
\newblock \emph{Neural Computation}, 30\penalty0 (1):\penalty0 271--291, 2017.

\bibitem[Du et~al.(2018)Du, Song, Hong, and Liu]{du2018learning}
Sizhen Du, Guojie Song, Haikun Hong, and Dong Liu.
\newblock Learning dynamic dependency network structure with time lag.
\newblock \emph{Science China Information Sciences}, 61\penalty0 (5):\penalty0
  1--16, 2018.

\bibitem[Du et~al.(2019)Du, Song, and Hong]{du2019collective}
Sizhen Du, Guojie Song, and Haikun Hong.
\newblock Collective causal inference with lag estimation.
\newblock \emph{Neurocomputing}, 323:\penalty0 299--310, 2019.

\bibitem[Zuo et~al.(2020)Zuo, Jiang, Li, Zhao, and Zha]{zuo2020transformer}
Simiao Zuo, Haoming Jiang, Zichong Li, Tuo Zhao, and Hongyuan Zha.
\newblock Transformer hawkes process.
\newblock In \emph{International conference on machine learning}, pages
  11692--11702. PMLR, 2020.

\bibitem[Zhang et~al.(2021)Zhang, Lipani, and Yilmaz]{zhang2021learning}
Qiang Zhang, Aldo Lipani, and Emine Yilmaz.
\newblock Learning neural point processes with latent graphs.
\newblock In \emph{Proceedings of the Web Conference 2021}, pages 1495--1505,
  2021.

\end{thebibliography}

\newpage
\appendix
\onecolumn

\section*{Appendix Overview}


In the following, we will provide supplementary materials to better illustrate our methods and experiments.

\begin{itemize} 

\item Appendix \ref{sec:related_work} describes our {\it related work}.

\item Appendix \ref{sec:identifiability} provides detailed {\it proof of Theorem 3}, i.e., the parameter identifiability of the time-lagged Hawkes processes.

\item Appendix \ref{sec:introduction_baselines} introduces the {\it baselines} we considered in the experiments.

\item Appendix \ref{sec:more_synthetic_data_experiment} presents the {\it supplementary} experiment results for our {\it synthetic} data, i.e., model parameter learning results for the sparse causal graph cases.

\item Appendix
\ref{sec:experiments_on_other_countries} provides the {\it supplementary} experiment results for our {\it real} data, i.e., the inferred lag posterior of government policies on confirmed Covid-19 cases in six countries.

\end{itemize}

\section{Related Work}
\label{sec:related_work}

\paragraph{Granger Causality and Temporal Point Process Model} Granger causality was initially used for studying the dependence structure for multivariate time series \citep{granger1969investigating, dahlhaus2003causality}, it has also been extended to multi-type event sequences \citep{didelez2008graphical}. Intuitively, for event sequence data, an event type is said to be (strongly) Granger causal for another event type if the inclusion of historical events of the former type leads to better predictions of future events of the latter type. Widely used existing point process model for inferring inter-type Granger causality is Hawkes process \citep{eichler2017graphical, xu2016learning, achab2017uncovering}, which assumes past events can only independently and additively excite the occurrence of future events according to a collection of pairwise kernel functions. Some works loosely referred to as neural point processes, has recently emerged in the literature such as~\citep{du2016recurrent, xiao2017modeling, mei2017neural, xiao2019learning}. {\it However, none of the existing temporal point process models explicitly consider the time lags in modeling the historical event effects}.

\paragraph{Time Lag Modeling} In terms of temporal models, an interesting yet less studied issue is the existence of time lags among different time series. That is, a past evidence would take some time to cause a future effect instead of an immediate response. \citet{gollob1987taking} took account of time lags in causal models. \citet{ye2015distinguishing} considered time lags and expanded convergent cross mapping, which is a statistical test for a cause-and-effect relationship between two variables. Some works on Granger causality make a strong assumption that every time point of the effect time series is influenced by a combination of other time series with a fixed time delay. However, the assumption of the fixed time delay does not always hold. To address this issue, \citet{amornbunchornvej2019variable} developed variable-lag Granger causality that relaxes the assumption of the fixed time delay and allows causes to influence effects with arbitrary time delays. \citet{absar2021discovering} addressed this time-lag limitation and proposed  a theoretically sound constraint-based algorithm which does not assume stationarity and works for both discrete and continuous time series.
In healthcare, the delayed effect of symptoms or treatment is worth the utmost vigilance of clinician. 
\cite{zhang2020adverse} focused on adverse drug reactions from electronic health records but ignore the potential time-varying confounding. \cite{wei2022granger} proposed a linear multivariate Hawkes process model used for Granger causal chain discovery for sepsis-associated derangements but also did not recognize delayed effect as vital factor. {\it We aim to model the delayed Granger causality based on temporal point process models, which is expected to wildly used for healthcare domain.}

\paragraph{Time Lag Estimation Methods} There are several works giving a shot to estimate the time lags from temporal data. \cite{dhurandhar2010learning} proposed to estimate the maximum lag existing among the temporal variables for grouped graphical Granger models. \cite{knapp1976generalized} proposed a maximum likelihood estimator for determining the time delay in causality analysis. A similar work proposed by \cite{sakurai2005braid} focused on estimating the time lag of stream causality discovery. \cite{shiller1973distributed} developed a distributed lag estimator from Bayesian priors regarding the smoothness of the lag curve. {\it These works only infer the lag from temporal data bu do not infer the causality at the same time.} \cite{zhou2017discovering} and \cite{du2017temporal, du2018learning} discovered dependencies based on time lag, {\it but they do not consider the relevance among target variables}. \cite{du2019collective} integrated both the collective causal relationships learning and time lag estimation simultaneously in one unified framework. {\it But in temporal point process setting, the flexible and effective time lag estimation method is still an untapped area.}

\section{Detailed Proof of Identifiability (Theorem 3)}
\label{sec:identifiability}
\vspace{-10pt}
In this section, we provide a detailed parameter identfiabiltiy proof. Let $\theta=\left(\theta_1, \ldots, \theta_U\right)$ be the model parameters with 
\begin{align}
\theta_u =\left( \mu_u, \,[\delta_{uu'}]_{u' = 1, \dots, U} \right) \in \Theta,
\end{align}
where $\Theta \subset \mathbb{R}_{+} \times \mathbb{R}_{+}^U$. 
Suppose we consider the exponential kennel $g(t) =\alpha \exp(-\beta(t-\delta))$ and we will have
\begin{align}
\theta_u =\left( \mu_u,\, [a_{uu'}]_{u' = 1, \dots, U},\,  [\beta_{uu'}]_{u' = 1, \dots, U}, \,[\delta_{uu'}]_{u' = 1, \dots, U} \right) \in \Theta,
\end{align}
where $\Theta \subset \mathbb{R}_{+} \times \mathbb{R}_{+}^U \times \mathbb{R}_{+}^U \times \mathbb{R}_{+}^U $. 

We prove the identifiability under the stationary assumption. Define $\lambda^{\text{delay}}(t\mid \mathcal{H}_t; \theta) = \{\lambda^{\text{delay}}_u(t \mid \mathcal{H}_t; \theta_u)\}_{u=1, \dots, U}$. Let $\mathcal{H}_t$ be a realization of the multivariate delayed Hawkes process. 
We are going to show that $\lambda_u^{\text{delay}}(t\mid \mathcal{H}_t; \theta_u) = \lambda_u^{\text{delay}}(t\mid \mathcal{H}_t; \tilde{\theta}_u)$ implies $\theta_u=\tilde{\theta}_u$. To this end, we will proceed the proof by showing that $\mu_u =\tilde{\mu}_u$, $\delta_{u} = \tilde{\delta}_{u}$, $\beta_{u} = \tilde{\beta}_{u}$, and $a_{u}= \tilde{a}_{u}$ in order, where we denote $\delta_u:=[\delta_{uu'}]_{u'\in [U]}$, $\beta_u:=[\beta_{uu'}]_{u'\in [U]}$ and $a_u:=[a_{uu'}]_{u'\in [U]}$.
For notational simplicity, we omit $\cH_t$ in the expression of the conditional intensity function.

\paragraph{(\it i)}\underline{Identifiability of $\mu_u$:} 

Let $t_{(1)}$ be the first event time in $\mathcal{H}_t$. For any $u$, since $\lambda^{\text{delay}}_u(t\mid \theta)= \lambda^{\text{delay}}_u(t\mid \tilde{\theta})$, we have $\int_0^{t_{(1)}}\lambda^{\text{delay}}_u(t\mid \theta)dt =\int_0^{t_{(1)}}\lambda^{\text{delay}}_u(t\mid \tilde{\theta})dt$. By definition, this leads to $(\mu_{u} -\tilde{\mu}_u)t_{(1)}=0$, and we conclude $\mu_{u}=\tilde{\mu}_u$. Note that this holds without any condition on $g$.

\paragraph{(\it ii)}\underline{Identifiability of $\delta_u$:}

To show this, we will exploit our assumptions on $g$ and the inter-arrival time.
Given the above condition, we first assume $\delta_u \neq \tilde{\delta}_u$. Since the inter-arrival time is continuously distributed, with probability one, the arrival times $\{t_k^u\}_{u,k}$ are all different from each other.
Let $S^u_{j}$ be the time of the $j$-th jump of the intensity function $\lambda^{\text{delay}}_u(t \mid \cH_t;\theta)$. By the assumption on $g_{uu'}$, every jump of the intensity function is excited by a unique arrival from some arrival process. 
Thus we have $
\{S_j^u\}_j = \bigcup_{u'\in[U]}\{t_k^{u'}+\delta_{uu'}\}_k =\bigcup_{u'\in[U]} \{t_k^{u'}+\tdelta_{uu'}\}_k.$
It follows that there exist $u',u''\in[U]$ such that the sets $\{t_k^{u'}+\delta_{uu'}\}_k$ and $\{t_k^{u''}+\tdelta_{uu''}\}_k$ have infinitely many common elements.
Note that $\delta_{uu'} \ne \tdelta_{uu''}$ since $\{t_k^u\}_{u,k}$ are all different with probability one.
It follows that there exists $k_1\ne k_2,\tilde{k}_1\ne \tilde{k}_2$ such that $t_{k_1}^{u'} - t_{\tilde{k}_1}^{u''} = t_{k_2}^{u'} - t_{\tilde{k}_2}^{u''} = \tdelta_{uu''}-\delta_{uu'}.$
But this occurs with probability zero, because the inter-arrival time is continuously distributed and the gaps between arrival times are all different with probability one.
Therefore, we have shown that with probability one, $\delta_u =\tilde{\delta}_u$.

\vspace{11pt}

For the exponential kernel, we will use the similar ideas as \cite{bonnet2022inference} to prove identifiability of $\beta_u$ and $a_u$, respectively.

\paragraph{(\it iii)}\underline{Identifiability of $\beta_u$:} 
Using \cite{bonnet2022inference} and our assumption on $\beta_u$, we can differentiate the intensity function and obtain $\beta_u \left(\lambda^{\text{delay}}_u(t\mid \theta) - \mu_u \right)=\tilde{\beta}_u  \left(\lambda^{\text{delay}}_u(t\mid \tilde{\theta}) - \mu_u \right) $. 
Since $\lambda^{\text{delay}}_u(t\mid \theta) - \mu_u >0$, we conclude that $\beta_u =\tilde{\beta}_u$.

\paragraph{(\it iv)}\underline{Identifiability of $a_u$:} 
Note that the event time in the process $N_{u'}(t-\delta_{uu'})$ are $ \{t^{u'}_k + \delta_{uu'}\}_k$.
Since $\lambda^{\text{delay}}_u(t\mid \theta)= \lambda^{\text{delay}}_u(t\mid \tilde{\theta})$, a.e., we first evaluate the intensity at time $\tau$ and get $\mu_u + \sum_{l=1}^U a_{ul} \sum_{t^{l}_k+ \delta_{ul}<\tau} e^{-\beta_u(\tau - t^{l}_k - \delta_{ul})} = \mu_u + \sum_{l=1}^U \tilde{a}_{ul} \sum_{t^{l}_k+ \delta_{ul}<\tau} e^{-\beta_u(\tau - t^{l}_k - \delta_{ul})}$. This leads to
\begin{align}
\sum_{l=1}^U ( a_{ul} - \tilde{a}_{ul}) A^l=0, \quad \text{with} \quad A^l=\sum_{t^{l}_k+ \delta_{ul}<\tau} e^{-\beta_u(\tau - t^{l}_k - \delta_{ul})}.
\label{eq:1}
\end{align} 
Similarly, we can evaluate the intensity at $\tau^+$, and obtain 
\begin{align}
\sum_{l=1}^U ( a_{ul} - \tilde{a}_{ul}) B^l=0, \quad \text{with} \quad B^l =\sum_{t^{l}_k+ \delta_{ul}<\tau^+} e^{-\beta_u(\tau^+ - t^{l}_k - \delta_{ul})}.
\label{eq:2}
\end{align}
Under our assumption on the event time, all events in this interval $[\tau, \tau^+)$ are from the process $N_{u'}(t-\delta_{uu'})$, so we have for all $l\neq u'$, $B^l = A^l e^{-\beta_u({\tau^+}-\tau)}$, and for $l = u'$, $B^{u'}= A^{u'}e^{-\beta_u(\tau^+ -\tau)}+ \sum_{\tau < t^{u'}_k + \delta_{uu'} <\tau^+}e^{-\beta_u(\tau^+ - t^{u'}_k - \delta_{uu'})}$. Plugging in Eq.~(\ref{eq:2}) and expanding $B^l$, we have
$$
e^{-\beta_u(\tau^+ -\tau)}\underbrace{\sum_{l=1}^U (a_{ul} - \tilde{a}_{ul})A^l}_{=0} + (a_{uu'}-\tilde{a}_{uu'}) \sum_{\tau < t^{u'}_k + \delta_{uu'} <\tau^+}e^{-\beta_u(\tau^+ - t^{u'}_k - \delta_{uu'})}=0.
$$
Given Eq.~(\ref{eq:1}), the above first term is zero. Then, the second term must be zero and this will lead to $a_{uu'}=\tilde{a}_{uu'}$. To prove $a_{uu}=\tilde{a}_{uu}$ we can let $\tau^+$ be the second event time of $u$ and adopt similar arguments. 
Thereby, we now have proved the theorem.
\endproof

\section{Introduction of Baselines in Experiments}
\label{sec:introduction_baselines}

We consider the following baselines through synthetic data experiments and healthcare data experiments to compare the event prediction with our proposed model:

\begin{itemize}

    

\item Transformer Hawkes Process (THP)~\citep{zuo2020transformer}: It is a concurrent self-attention based point process model with additional structural knowledge.
\item Recurrent Marked Temporal Point Processes (RMTPP) ~\citep{du2016recurrent}: It uses RNN to learn a representation of the past events and time intervals.
\item ERPP~\citep{xiao2017modeling}: It uses two RNN models for the background of intensity function and the history-dependent intensity part. 
\item Granger Causal Hawkes (GCH)~\citep{xu2016learning}: A multivariate Hawkes model with sparse-group-lasso and pairwise similarity constriants. It uses an EM learning algorithm with the basis function of Hawkes process selected adaptively. 
\item LG-NPP algorithm~\citep{zhang2021learning}: It applies self-attention to embed the event sequence data. It uses a latent random graph to model the relationship between different event sequences and proposes a bilevel programming algorithm to uncover the latent graph and embedding parameters.
\item GM-NLF algorithm~\citep{eichler2017graphical}: It is a  multivariate Hawkes Processes with nonparameteric intensity function.

\item Variational Inference (VI): Without using an amortized neural network to approximate the posterior of the time lag by conditioning on the observation, we approximate the posterior using simple factorized parametric distributions. All the  posterior parameters and model parameters are optimized via ELBO.

\item MLE: We take the parameters of time lag distribution as learnable parameters and learn these parameters by directly optimizing the likelihood.

\end{itemize}

\section{More Synthetic Data Experiments}
\label{sec:more_synthetic_data_experiment}

\paragraph{Inference}

We consider three posterior parametric distributions for the latent time lags, including normal distribution, log-normal
distribution (heavy-tailed) and exponential distribution. In practice, which model will be the best is unknown beforehand. To compare more patterns, we also assume that the Granger causal graph with a sparsity structure is known and given as prior knowledge. We further consider three given sparsity structures: upper triangle, lower triangle, and four corners respectively.

\begin{figure}[ht]
\centering 
\includegraphics[width=0.7\textwidth]{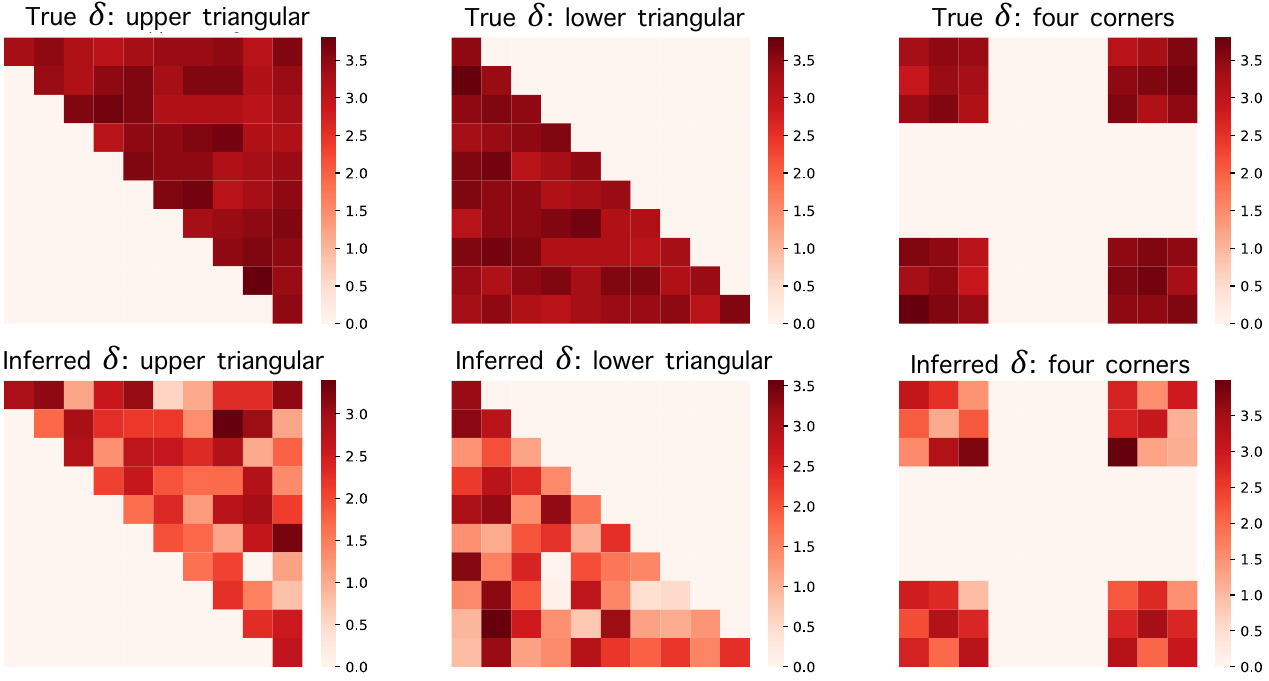} 
\caption{Inferred time lags: mean parameter of normal distribution. Results using VAE with known pattern before training.}
\label{fig:VAE_normal_pattern_appendix}
\end{figure}

\begin{figure}[ht]
\centering 
\includegraphics[width=0.7\textwidth]{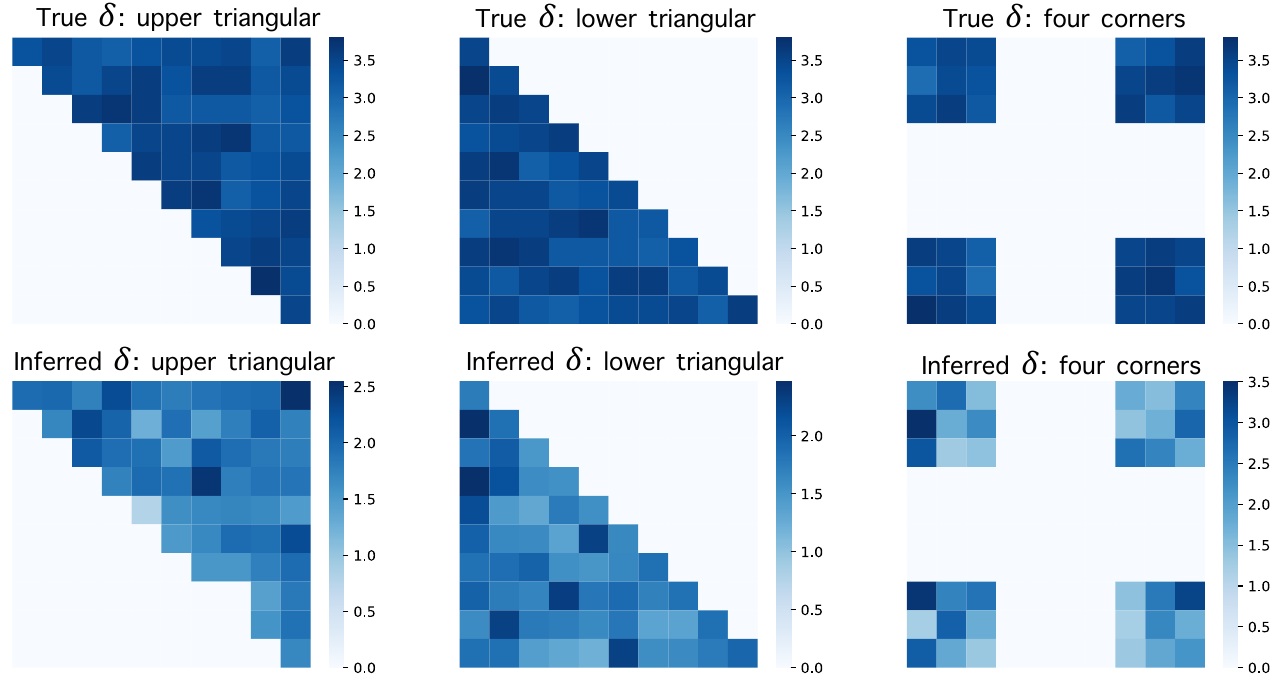} 
\caption{Inferred time lags: rate parameter of exponential distribution. Results using VAE with known pattern before training.}
\label{fig:VAE_exponential_pattern}
\end{figure}

\begin{figure}[ht]
\centering 
\includegraphics[width=0.7\textwidth]{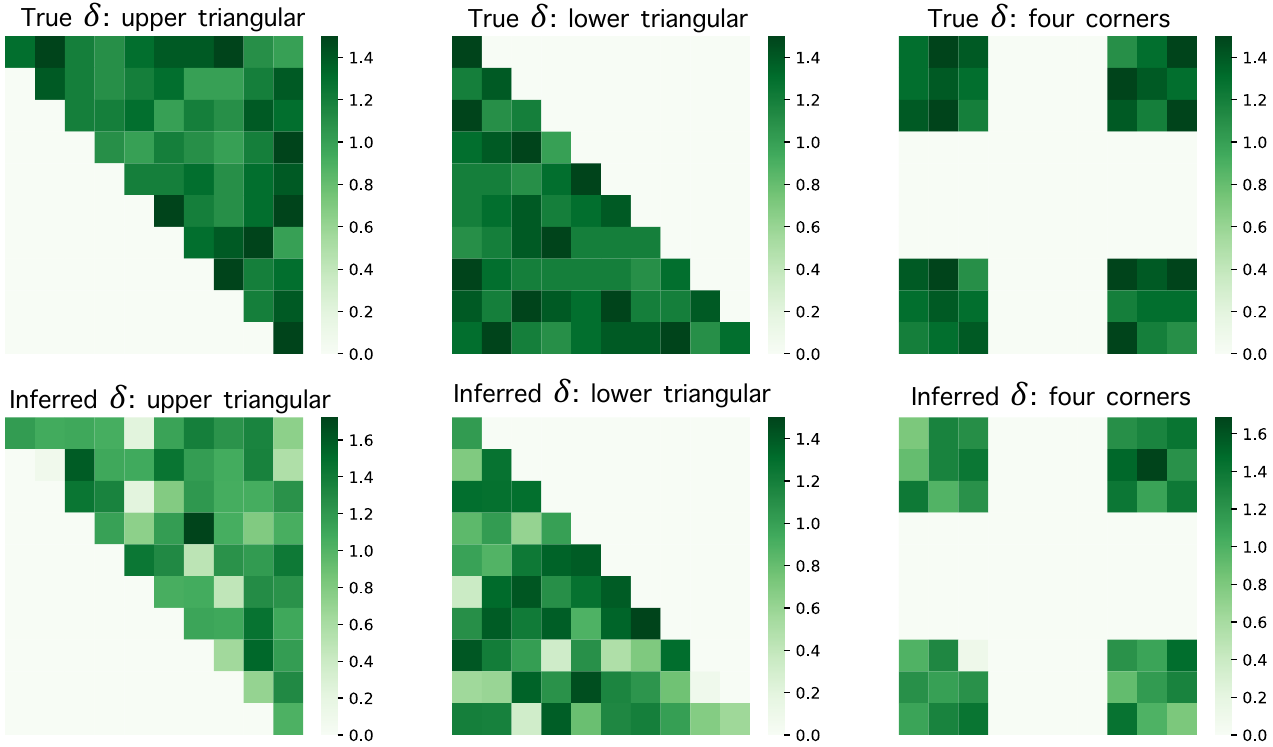} 
\caption{Inferred time lags: mean parameter of log-normal distribution. Results using VAE with known pattern before training.}
\label{fig:VAE_lognormal_pattern}
\end{figure}

We compared the {\it inferred time lags} (posterior mean) with the true time lags in Figs.~\ref{fig:VAE_normal_pattern_appendix},~\ref{fig:VAE_exponential_pattern}, and~\ref{fig:VAE_lognormal_pattern}. From the results, we conclude that our method achieves satisfactory performance in inferring the latent time lags. Note that these figures only shown the mean parameter of normal distribution, the rate parameter of exponential distribution, and the mean parameter of the log-normal distribution, although we can infer the entire posterior distributions.

\vspace{-10pt}

\section{Experiments on Other Countries}
\label{sec:experiments_on_other_countries}

\begin{figure}[H]
\centering 
\includegraphics[width=0.7\textwidth]{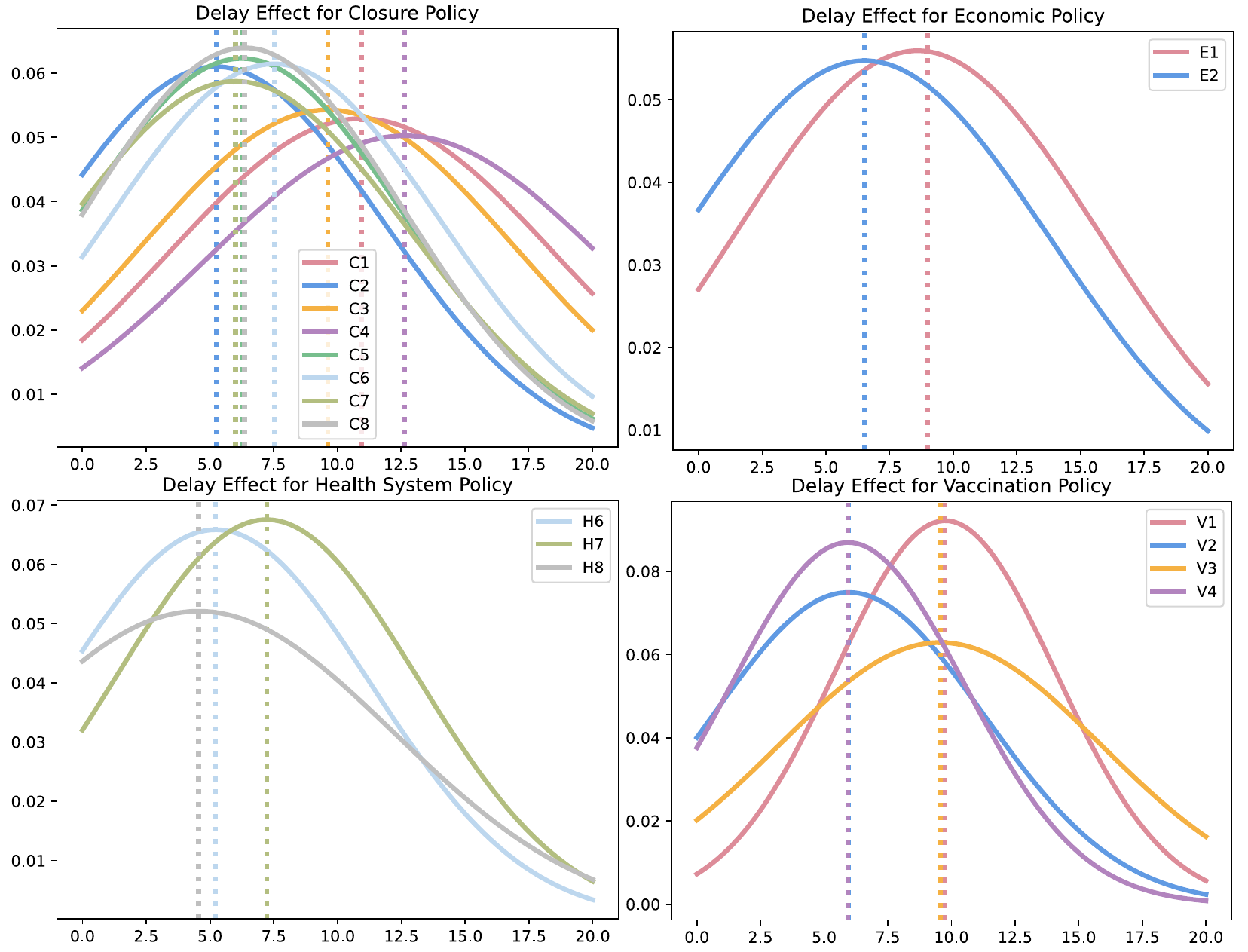} 
\caption{Inferred lag posterior of government policy on confirmed cases using Covid-19 Policy Tracker data in Australia.}
\label{fig:covid_delayed_effect_Australia}
\end{figure}

\begin{figure}[H]
\centering 
\includegraphics[width=0.7\textwidth]{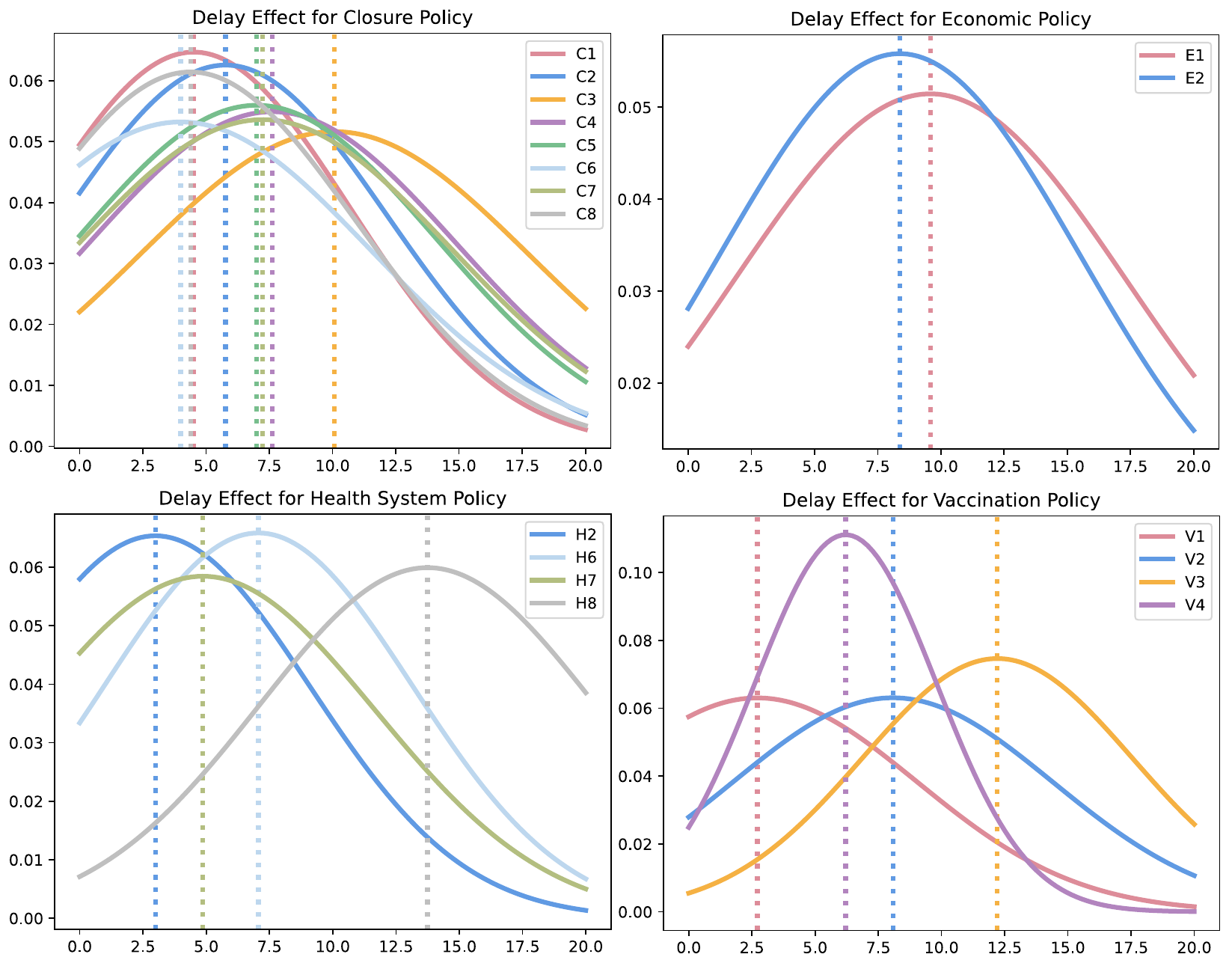} 
\caption{Inferred lag posterior of government policy on confirmed cases using Covid-19 Policy Tracker data in France.}
\label{fig:covid_delayed_effect_France}
\end{figure}

\begin{figure}[H]
\centering 
\includegraphics[width=0.7\textwidth]{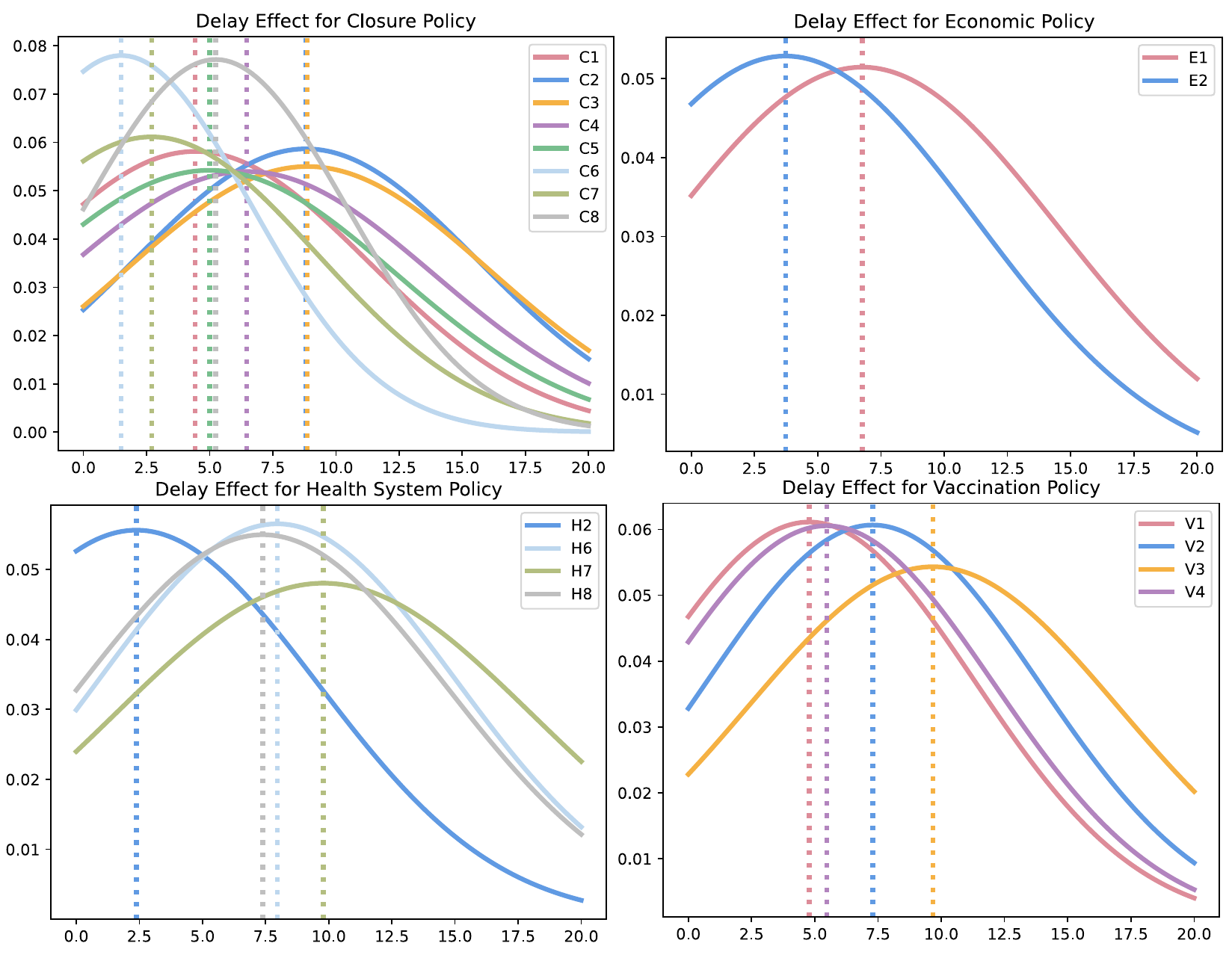} 
\caption{Inferred lag posterior of government policy on confirmed cases using Covid-19 Policy Tracker data in Italy.}
\label{fig:covid_delayed_effect_Italy}
\end{figure}

\begin{figure}[H]
\centering 
\includegraphics[width=0.7\textwidth]{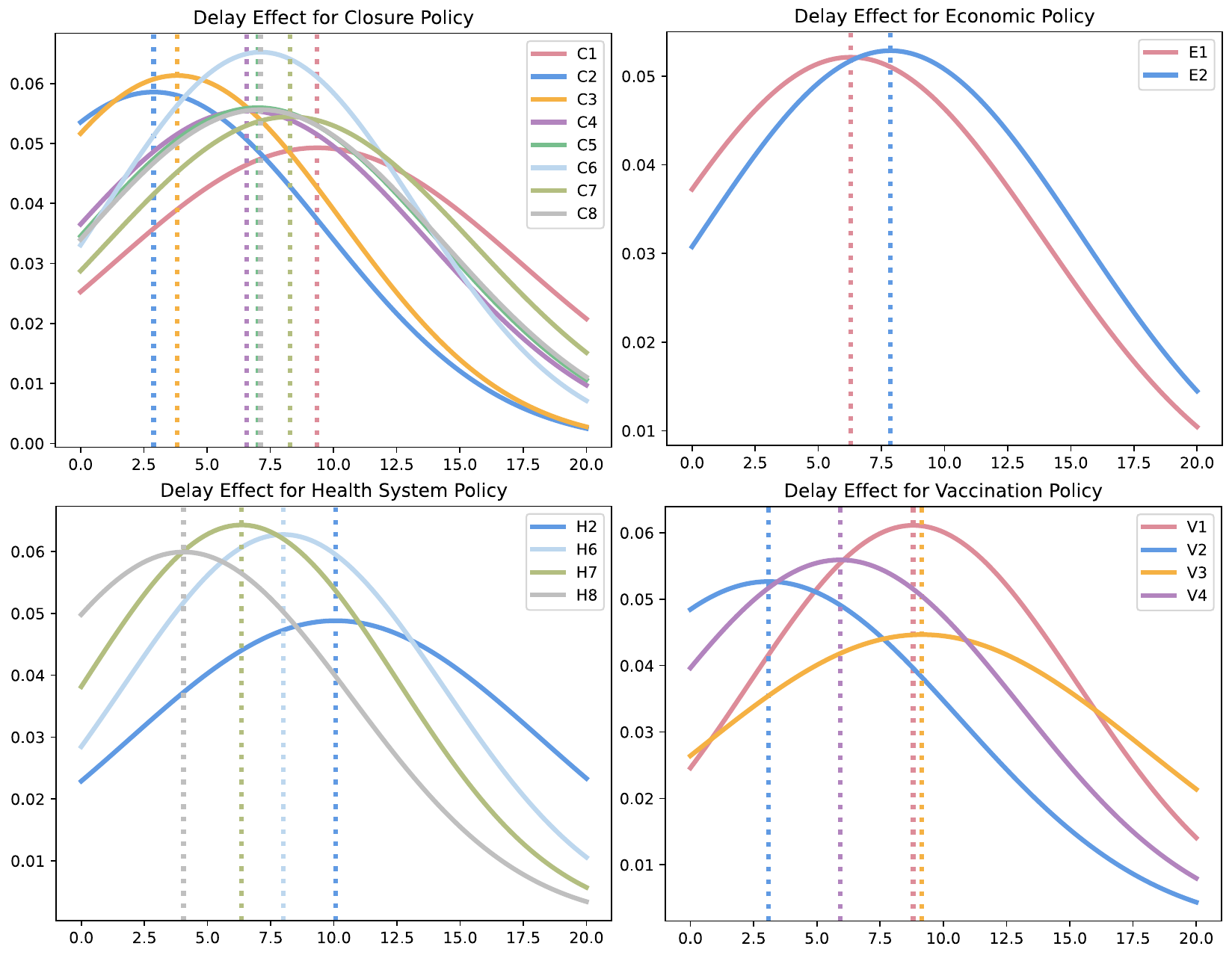} 
\caption{Inferred lag posterior of government policy on confirmed cases using Covid-19 Policy Tracker data in South Korea.}
\label{fig:covid_delayed_effect_South_Korea}
\end{figure}

\begin{figure}[H]
\centering 
\includegraphics[width=0.7\textwidth]{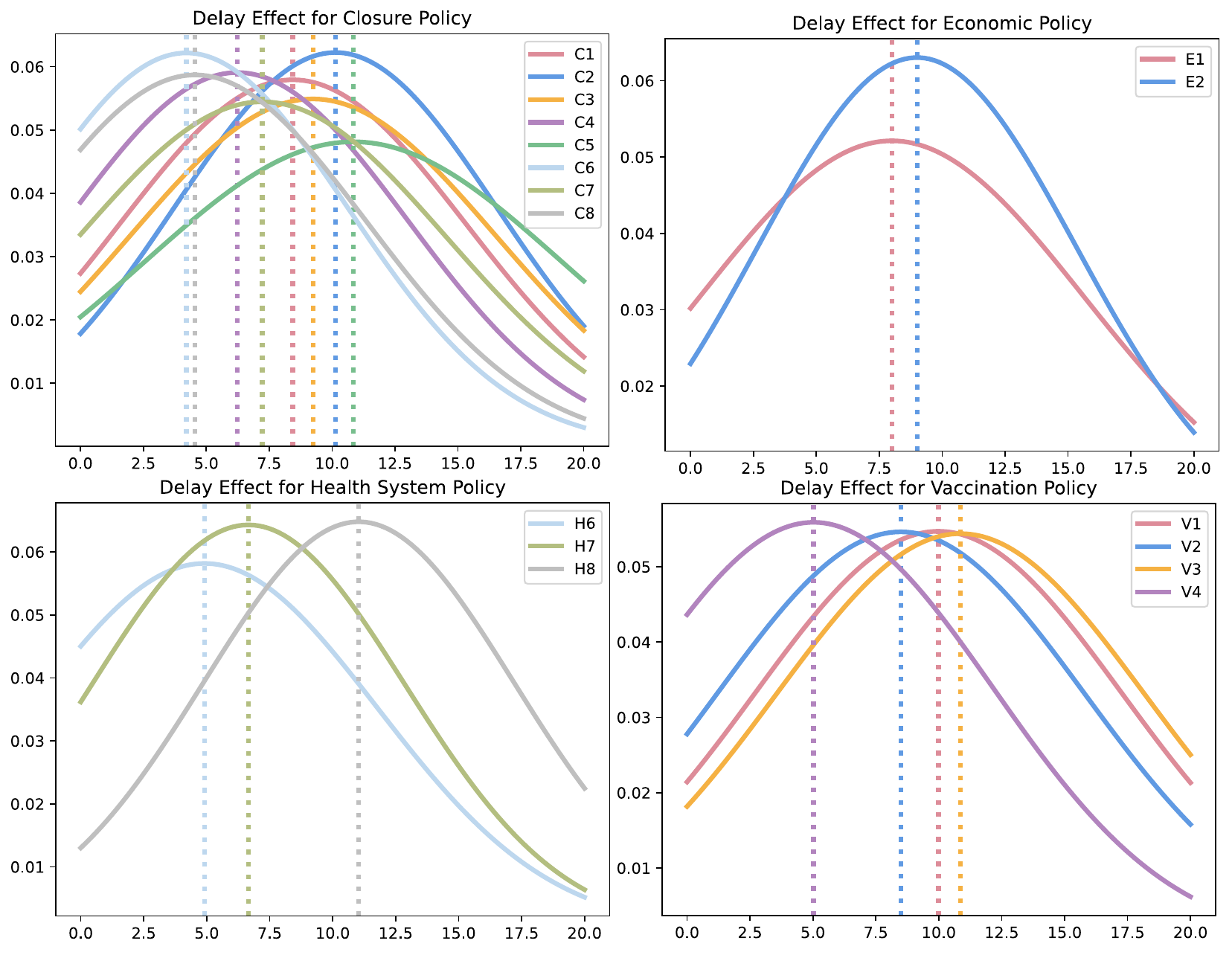} 
\caption{Inferred lag posterior of government policy on confirmed cases using Covid-19 Policy Tracker data in United Kingdom.}
\label{fig:covid_delayed_effect_United_Kingdom}
\end{figure}

\begin{figure}[H]
\centering 
\includegraphics[width=0.7\textwidth]{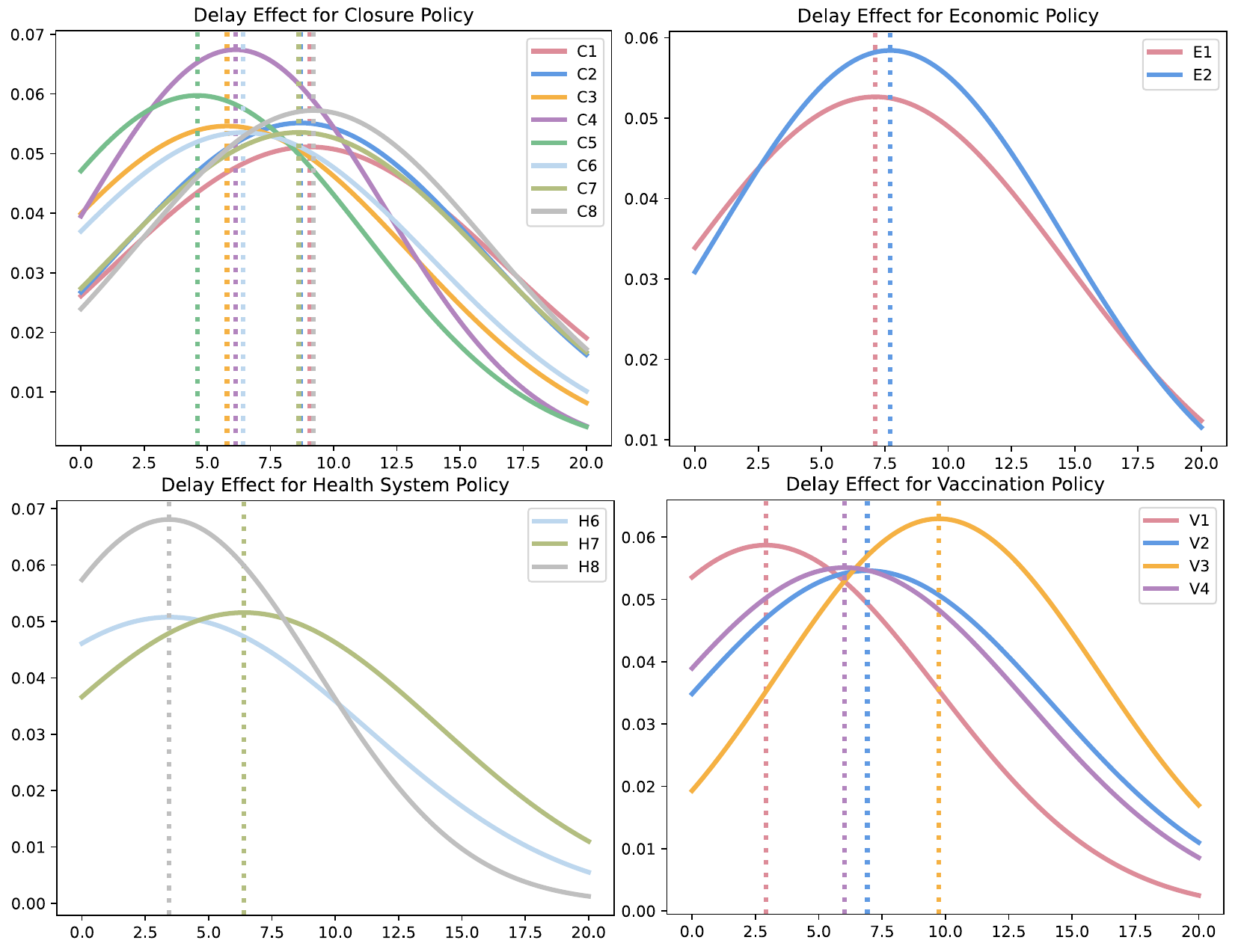} 
\caption{Inferred lag posterior of government policy on confirmed cases using Covid-19 Policy Tracker data in United States.}
\label{fig:covid_delayed_effect_United_States}
\end{figure}

The results of other six nations (Australia, France, Italy, South Korea, United Kingdom, United States) are summarized in Tab.\ref{table_sumarize 6 nations results} followed by an explanatory Tab.\ref{Policy code description}. In Tab.\ref{table_sumarize 6 nations results}, we tick the policy for nations if it probably took effect within one week. For intuitive presentation, results are also plotted in Fig. \ref{fig:covid_delayed_effect_Australia}, Fig. \ref{fig:covid_delayed_effect_France}, Fig. \ref{fig:covid_delayed_effect_Italy}, Fig. \ref{fig:covid_delayed_effect_South_Korea}, Fig. \ref{fig:covid_delayed_effect_United_Kingdom}, and Fig. \ref{fig:covid_delayed_effect_United_States}.

Mandatory vaccination (V4) policy is efficient across all six nations, which could be summarized into two reasons. First is that policy is mandatory which would not have problem of adherence. Another reason relates to the fact that vaccination is standardised which could not vary between nations. For the remaining policies, they could be efficient in some countries, but have long time delay in others. For example, staying at home (C6) worked probably around 1.5 days and its variance is 5 days in Italy, but needed 7.5 days to take effect in Australia. This phenomenon could be caused by the vast difference in the territories of these two countries, which directly impact on population density. Moreover, medical experts suggest that wearing mask is an effective way to prevent the spread of virus, but facial coverings (H6) usually took 8 days delay to control epidemic in Italy, indicating that low adherence of Italian to H6. Low adherence to H6 also could explain fast response of C6. Moreover, in United Kingdom, the short time delay (4.5 days) of policy restrictions of international travel (C8) suggests that the UK, as the first country to introduce herd immunity, is more vulnerable to exotic strains. Our proposed model could perform as a powerful tool for government officials to know their national conditions and take more efficient decisions to prevent epidemic.

\begin{table*}[t]
\centering
\begin{tabular}{c|c|c|c|c|c|c|c|c|c|c|c|c|c|c|c|c|c|c}
\hline
\multirow{3}{*}{\textbf{Nations}} & \multicolumn{18}{c}{\textbf{Policies}}\\
\cline{2-19}
 &\multicolumn{8}{c|}{Containment \& closure}&\multicolumn{4}{c|}{Health system}&\multicolumn{4}{c|}{Vaccination}&\multicolumn{2}{c}{Economic}\\
\cline{2-19}
 & C1 & C2 & C3 & C4 & C5 & C6 & C7 & C8 & H2  & H6 & H7 & H8 & V1 & V2 & V3 & V4 & E1 & E2 \\
\hline
\textbf{AUS} & & \checkmark &  & &\checkmark  & &\checkmark & \checkmark & &\checkmark &  &\checkmark & &\checkmark  & &\checkmark &  &\checkmark \\
\hline
\textbf{SK} & & \checkmark & \checkmark &\checkmark &  & & &  & & & \checkmark & \checkmark& & \checkmark & & \checkmark& \checkmark &\\
\hline
\textbf{UK} & &  &  &\checkmark &  & \checkmark& & \checkmark & & \checkmark & \checkmark & & &  & & \checkmark&  &\\
\hline
\textbf{US} & &  & \checkmark &\checkmark & \checkmark &\checkmark & &  & &\checkmark & \checkmark &\checkmark & \checkmark& \checkmark & & \checkmark&  &\\
\hline
\textbf{Italy} &\checkmark &  &  &\checkmark &\checkmark  &\checkmark & \checkmark& \checkmark &\checkmark & &  & &\checkmark &  & &\checkmark & \checkmark &\checkmark\\
\hline
\textbf{France} & \checkmark& \checkmark &  & & \checkmark &\checkmark & & \checkmark & \checkmark&\checkmark & \checkmark & & \checkmark&  & &\checkmark &  &\\
\hline
\multicolumn{19}{l}{{\tiny \textit{Note}: \textbf{AUS} for Australia;\  \textbf{SK} for South Korea; \textbf{UK}: United Kingdom; \textbf{US}: United States.}}\\
\hline
\end{tabular}
\caption{Policies which would probably take effects within one week for each nation.}
\label{table_sumarize 6 nations results}
\end{table*}
\begin{table*}[t]
\centering
\begin{tabular}{l|c|l}
\hline
\multicolumn{1}{c|}{\textbf{Category}} & \textbf{Code} & \multicolumn{1}{c}{\textbf{Explain}} \\
\hline
\multirow{8}{*}{\textbf{Containment \& closure policies}} & C1 & School closing.\\
\cline{2-3}
 & C2 & Workplace closing.\\
\cline{2-3}
 & C3 & Cancel public events.\\
\cline{2-3}
 & C4 & Restrictions on gatherings.\\
\cline{2-3}
 & C5 & Close public transport.\\
\cline{2-3}
 & C6 & Stay at home requirements.\\
\cline{2-3}
 & C7 & Restrictions on internal movement.\\
\cline{2-3}
 & C8 & International travel controls.\\
\hline
\multirow{8}{*}{\textbf{Health system policies}} & H1 & Public information campaigns.\\
\cline{2-3}
 & H2 & Testing policy.\\
\cline{2-3}
 & H3 & Contact tracing.\\
\cline{2-3}
 & H4 & Emergency investment in healthcare.\\
\cline{2-3}
 & H5 & Investment in vaccines.\\
\cline{2-3}
 & H6 & Facial coverings.\\
\cline{2-3}
 & H7 & Vaccination policy.\\
\cline{2-3}
 & H8 & Protection of elderly people.\\
\hline
\multirow{4}{*}{\textbf{Vaccination policies}} & V1 & Vaccine prioritisation.\\
\cline{2-3}
 & V2 & Vaccine eligibility/availability.\\
\cline{2-3}
 & V3 & Vaccine financial support.\\
\cline{2-3}
 & V4 & Mandatory Vaccination.\\
\hline
\multirow{4}{*}{\textbf{Economic policies}} & E1 & Income support.\\
\cline{2-3}
 & E2 & Debt/contract relief.\\
 \cline{2-3}
 & E3 & Fiscal measures.\\
 \cline{2-3}
 & E4 & International support.\\
 \hline
\textbf{Miscellaneous policies} & M1 & Record other policy announcements.\\
\hline
\end{tabular}
\caption{Policies description of each code.}
\label{Policy code description}
\end{table*}

\end{document}